\documentclass[fleqn,10pt]{wlscirep}
\usepackage[utf8]{inputenc}
\usepackage[T1]{fontenc}
\usepackage[mathscr]{euscript}
\usepackage{bm}% bold math
\usepackage{etoolbox}
\usepackage{amsmath}
\usepackage{amssymb}
\usepackage{xcolor}
\usepackage{nicefrac}
\usepackage{float}
\usepackage{musicography}
\usepackage{verbatim}

\DeclareMathOperator{\supp}{supp}
\DeclareMathOperator{\diag}{diag}

\title{Distinguishing indistinguishable attractors: Unsupervised anomaly detection with reservoir computers}

\author[1]{Davide Prosperino}
\author[1]{Haochun Ma}
\author[2,*]{Christoph Räth}
\affil[1]{Ludwig-Maximilians-Universität München, Faculty of Physics, Munich, 80539, Germany}
\affil[2]{Deutsches Zentrum für Luft- und Raumfahrt (DLR), Institute for Frontier Materials on Earth and in Space, Cologne, 51170, Germany}

\affil[*]{christoph.raeth@dlr.de}


\begin{abstract}
Detecting when a nonlinear dynamical system departs from its normal regime is a recurring problem across the sciences, from cardiology to climate and energy systems.
We show that a very simple Kolmogorov--Smirnov test on the output weights of a reservoir computer is highly sensitive to regime changes in nonlinear dynamical systems, including those invisible to both classical nonlinear measures and modern deep-learning detectors.
The core idea of our algorithm is to treat the readout layer of a reservoir computer
as a representation of the input dynamics.
Since the input mapping and the reservoir itself are random and fixed, the trained output weights are the only object encoding the system at hand.
We summarize this fingerprint by the empirical cumulative distribution function of the readout weights and compare it to a reference band built from the training data.
This unsupervised, online detector distinguishes two visually indistinguishable butterfly-shaped attractors, resolves parameter drifts seven times smaller than the strongest deep-learning baseline, flags noise four orders of magnitude below the signal, and identifies ventricular flutter in a clinical ECG recording.
More broadly, we aim to establish a perspective on reservoir computers in which the trained output weights are treated as a representation of the learned system in their own right, rather than merely as a means to forecasting.
\end{abstract}

\begin{document}

\flushbottom
\maketitle

\thispagestyle{empty}

\section*{Introduction}
Detecting anomalous behavior in complex dynamical systems is a recurring problem across the sciences, from cardiology to energy demand and food security.
Reservoir computing (RC) has emerged as a powerful tool for modeling such systems\cite{Shahi2021, Brucke2024, Herteux2024}.
However, the majority of RC-based algorithms, from anomaly detection\cite{Kato2024} to causality analysis\cite{Wang2022, Zhao2024}, rely on the reservoir's forecasting output.
In contrast, a few recent works probe other, more intrinsic properties of the trained reservoir computer\cite{Banerjee2019, Cao2025, Tamura2026}.
In a similar spirit, in this work we introduce an anomaly-detection algorithm that operates purely on the learned output weights.
More broadly, we want to shift the perspective on reservoir computers away from their usual role as forecasting systems and emphasize the descriptive power of the trained output weights as a characterization of the learned system.

We show that a simple Kolmogorov--Smirnov test on the distribution of output weights is sensitive enough to distinguish visually indistinguishable attractors, to flag barely perceptible parametric drift, and to detect the onset of weak observational noise.
Tasks, where deep-learning benchmarks fail.
Finally, we demonstrate that the same algorithm, applied unchanged, is able to identify a pathological episode in a real-world ECG recording.

The simple, core idea of our algorithm is to treat the readout layer of a reservoir computer as a representation of the input dynamics.
Since the input mapping and the reservoir are both random and fixed in a classical RC setup, the readout is the only component fitted to the data, and therefore the only object containing parameters that encodes the dynamics.

\begin{figure}[ht]
\centering
\includegraphics[width=\textwidth]{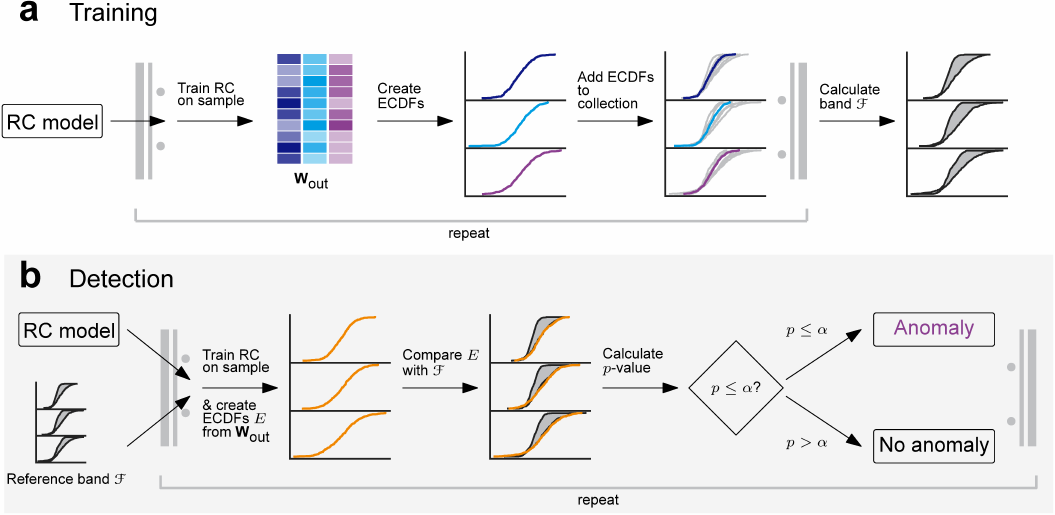}
\caption{\textbf{Schematic version of our algorithm.}
Musical repetition signs bracket the per-sample loop, iterated over training samples in \textbf{a} and over evaluation samples in \textbf{b}.
Panel \textbf{a} shows the training routine:
For each sample, we train the RC model, extract its readout layer, and convert it into ECDFs. Collecting these ECDFs across all training samples yields the reference band $\mathscr{F}$.
Panel \textbf{b} shows the detection routine, which takes both the RC model and the reference band $\mathscr{F}$ as input.
For each new sample, we again train the RC model and form an ECDF $E$ from its readout layer, then compute a $p$-value by comparing $E$ against $\mathscr{F}$.
A $p$-value below the threshold $\alpha$ flags the sample as anomalous.
In principle, the detection loop runs indefinitely.}
\label{fig:schema}
\end{figure}

For our algorithm, we utilize this by training the reservoir on a sliding window of the data and recording, for each window, the empirical cumulative distribution function (ECDF) of the resulting output weights.
The collection of these ECDFs spans a reference band $\mathscr{F}$ that captures the natural variability of the fingerprint under the training dynamics.
Fig. \ref{fig:schema}\textbf{a} illustrates this collection in the training phase.
During the detection phase, shown in Fig. \ref{fig:schema}\textbf{b}, we train the reservoir computer on the new sample to retrieve its fingerprint.
A Kolmogorov--Smirnov test then quantifies, under the assumption that the new fingerprint comes from the reference band $\mathscr{F}$, how likely we are to see this much deviation by chance.
The result is a $p$-value.
A $p$-value below the threshold $\alpha$ flags the sample as anomalous.

Our algorithm presupposes a reservoir that adequately models the system at hand, though we find it to be robust to the exact choice of hyperparameters beyond this requirement.
We refer to existing methodological guides for the underlying configuration\cite{Lukosevicius2012, Haluszczynski2019, Racca2021}.

Throughout this article we use the point-wise $F_1$ score as a measure for quality, as it balances precision and recall in a single number and thereby penalizes both missed anomalies and false alarms.
It is calculated as the harmonic mean between precision $P$ and recall $R$ as
\begin{equation}
    F_1 = \dfrac{2}{P^{-1} + R^{-1}} = \dfrac{2\,\textrm{True Positive}}{2\,\textrm{True Positive} + \textrm{False Positive}+\textrm{False Negative}}\;\;\;.
\end{equation}
We do not apply the point-adjustment protocol common in parts of the time-series anomaly-detection literature, as it tends to inflate reported numbers\cite{Kim2022}.

We benchmark our algorithm on synthetic systems against three representative anomaly-detection methods identified by the recent review of Mejri \textit{et al.}\cite{Mejri2024}: the Graph Deviation Network (GDN)\cite{Deng2021}, the USAD model\cite{Audibert2020}, and a modern online variant of ARIMA\cite{Kozitsin2021}.
The first two are deep-learning models, and ARIMA is a classical statistical baseline.
Beyond detection quality, this comparison also exposes a substantial efficiency gap: Our RC-based detector trains in seconds on a CPU, whereas the deep baselines require minutes to tens of minutes and benefit substantially from GPU acceleration.
Even relative to the deep baselines running on a GPU, our CPU-only detector trains substantially faster.
A margin that matters whenever detection must run online, on edge hardware, or under tight energy budgets.

Taken together, our results position the readout-weight distribution of reservoir computers as a compact fingerprint of nonlinear dynamics that captures regime changes invisible to both classical measures and deep-learning detectors.
The remainder of this article presents these results in detail, followed by the methodological background in the Methods section.

\section*{Results}
We evaluate our algorithm on four scenarios, spanning structural changes in the governing equations, parametric drift, additive observational noise, and a transition between physiological regimes in clinical electrocardiogram data.
The first three are based on the Lorenz system\cite{Lorenz1963} and test complementary aspects of the detector: sensitivity to a change of governing equations under a visually invariant attractor, sensitivity to changing parameters, and robustness to measurement noise.
The fourth scenario applies the same algorithm to a real-world nonlinear system whose true equations of motion are unknown.
The synthetic tests are benchmarked against three models selected following the recent review by Mejri \textit{et al.}\cite{Mejri2024}: the predictive Graph Deviation Network (GDN)\cite{Deng2021}, the reconstruction-based USAD model\cite{Audibert2020}, and a modern online variant of ARIMA\cite{Kozitsin2021}.
Throughout, we focus on the $F_1$ score as evaluation metric.

Across all experiments, our proposed RC architecture trains orders of magnitude faster than the deep-learning benchmarks.
The computational cost of all detectors is summarized in Tab. \ref{tab:compute}.
Even on a CPU, our RC-based detector trains substantially faster than the deep baselines do on a GPU, by roughly an order of magnitude relative to GDN and nearly two relative to USAD.
Among all methods, only the classical ARIMA baseline is cheaper to fit.
In our framework, training and online updating are the same operation:
Each new sample triggers a full retraining of the readout, so the entire reference collection $\mathscr{F}$ is assembled at roughly the same per-step cost as a single online update.
The trade-off shifts at inference time.
Once trained, the deep baselines predict in fractions of a millisecond, and ARIMA faster still, whereas our detector pays the full training cost per sample.
For the settings considered here, sub-second updates are well within the operational budget, and the absence of any GPU requirement is the more relevant practical constraint.

\begin{table}[ht]
\caption{\label{tab:compute}\textbf{Computational cost of the detection methods.}
Total training times and per-step inference times are reported in seconds and milliseconds, respectively.
For the deep baselines GDN \& USAD, CPU values are given first with the GPU times in parentheses.
All values refer to the dynamical-system switching experiment.
They should be read as rough estimates, indicating relative cost, rather than precise measurements.}
\centering
\begin{tabular}{|l|c|c|}
\hline
Model & Total training time [s] (GPU)& Inference time per step [ms] (GPU)\\
\hline
RC & $14$ & $150$\\
NGRC & $3$ & $40$\\
\hline
GDN & $680$ ($60$) & $0.3$ ($0.1$) \\
USAD & $1\,120$ ($330$) & $0.1$ ($0.1$)\\
ARIMA & $3$ & $7$\\
\hline
\end{tabular}
\end{table}

\subsection*{Switching dynamical systems}
\begin{figure}[ht]
\centering
\includegraphics[width=\linewidth]{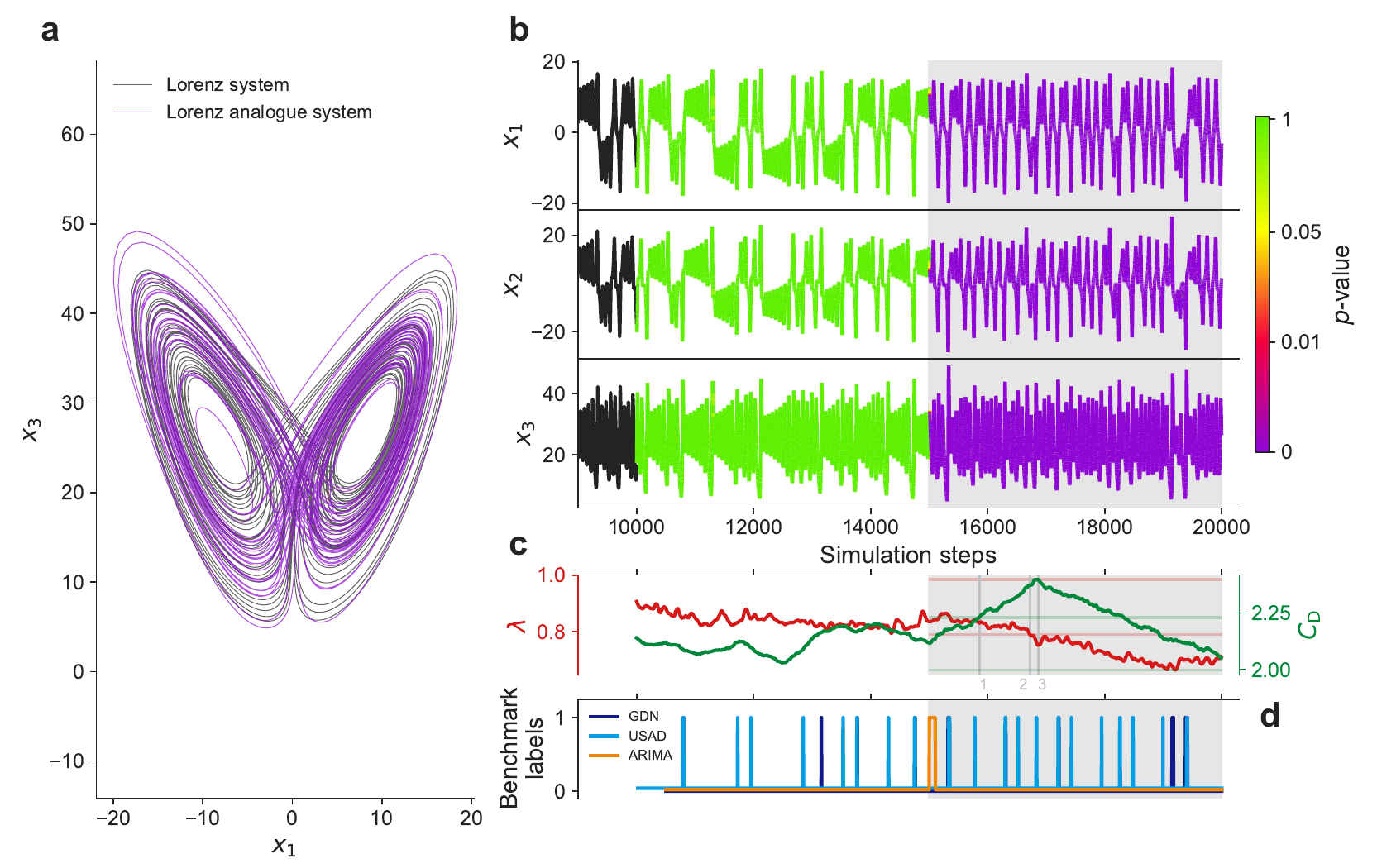}
\caption{\textbf{Distinguishing visually indistinguishable attractors.}
Subfigure \textbf{a} shows the projection of the attractors of the classical Lorenz system in black and the anomalous Lorenz analogue system in purple.
Visually, the attractors are indistinguishable from each other.
Subfigure \textbf{b} shows the detection results on a test trajectory in which the dynamics switch from the Lorenz system (white background) to the Lorenz analogue system (grey background) at simulation step $15\,000$.
The color in those panels encodes the $p$-value of our RC detection model for each coordinate separately, while black indicates training data.
A low $p$-value indicates anomalous behavior.
Subfigure \textbf{c} shows the rolling largest Lyapunov exponent $\lambda$ in red, and the rolling correlation dimension $C_\textrm{D}$ in green.
The faint lines indicate the bounds of the three-sigma interval for each measure. 
Lastly, subfigure \textbf{d} shows the binary labels for the three anomaly-detection benchmark methods.
} %350 words max
\label{fig:lorenz-lorenzanalogue}
\end{figure}

We evaluate the sensitivity of our algorithm to structural changes in the governing equations of a dynamical system.
While a change to a qualitatively different system---\textit{e.g.} from the Lorenz system to the Halvorsen system---may be trivial to detect using simple statistical analysis, we focus on changes to governing equations which produce visually indistinguishable trajectories and result in the same attractor.
Concretely, we study the switch from the classical Lorenz system to an alternative realization of the Lorenz attractor\cite{Prosperino2025} (see Eqs. \ref{equ:library-order-2}), hereafter referred to as the Lorenz analogue system.
The visually indistinguishable trajectories are shown in Fig. \ref{fig:lorenz-lorenzanalogue}\textbf{a}, with black depicting the Lorenz attractor and purple the Lorenz analogue attractor.
Despite the similarity of the resulting trajectories, the two systems are not related by parameter variation, coordinate transformation, or perturbation.

Following the classification of anomalies by Lai \textit{et al.}\cite{Lai2021}, a change in a dynamical system best fits the shapelet or seasonal type, for which the review by Mejri \textit{et al.}\cite{Mejri2024} identifies ARIMA-based methods, the Graph Deviation Network (GDN)\cite{Deng2021}, and USAD\cite{Audibert2020} as the best-performing approaches.
We adopt these three as benchmarks and defer their configuration to the discussion of Fig. \ref{fig:lorenz-lorenzanalogue}\textbf{d}.

The complete training dataset consists of $10\,000$ steps of the Lorenz system, and the test dataset consists of $5\,000$ steps of the Lorenz system, after which we switch to the Lorenz analogue system for another $5\,000$ steps, marking the anomaly.
The last $1\,000$ steps of the training data and the test data are shown in the first three panels of Fig. \ref{fig:lorenz-lorenzanalogue}\textbf{b}.

For our RC detection model, we fix our reservoir to a scale-free network with $d=500$ nodes and scale it to a target spectral radius of $\rho_{\textrm{t}}=0.1$.
We build our collection of reference ECDFs $\mathscr{F}$ using a sliding window on the training data with a step size of $2$ and collect a total of $50$ ECDFs for each coordinate.
The collection $\mathscr{F}$, therefore, contains the combined information of $5\,100$ data points.
Each ECDF is obtained by training the model using $5\,000$ data points of the unscaled training data and using a regularization parameter of $\beta=10^{-6}$ for the ridge regression.

We present our results in Fig. \ref{fig:lorenz-lorenzanalogue}\textbf{b}:
The $p$-value drops the earliest for the $x_3$-coordinate, where it drops below $1 \%$ after including only $2$ data points of the anomalous system.
For the $x_1$ coordinate it takes $11$ data points to drop below the threshold, and for $x_2$ it takes $94$ data points.
After the $p$-value falls, it stays at numerically $0$ during all anomalous data points.
Our RC based algorithm is therefore capable of instantly detecting the switch to a new dynamic.

We benchmark our results against two families of baselines: classical nonlinear-systems measures, namely the largest Lyapunov exponent $\lambda$ (hereafter simply the Lyapunov exponent) and the correlation dimension $C_\textrm{D}$, and modern anomaly-detection models.
The classical measures are shown in Fig. \ref{fig:lorenz-lorenzanalogue}\textbf{c}.

In Ref. \citeonline{Prosperino2025} we found the Lyapunov exponent of the Lorenz analogue system to be smaller than the Lyapunov exponent of the original Lorenz system.
In principle, one could therefore distinguish the systems by their Lyapunov exponents.
In practice, however, as we show below, this requires a substantial number of anomalous data points to accumulate before the estimate has converged enough to separate the two regimes.
We test this by calculating the Lyapunov exponent and correlation dimension on a rolling window of $15\,000$ and $5\,000$ steps, respectively, as we found the measures to converge after that many steps for the original Lorenz system.
The Lyapunov exponent is computed using the method of Rosenstein \textit{et al.}\cite{Rosenstein1993}, and the correlation dimension using the Grassberger--Procaccia algorithm\cite{Grassberger1983}.

We notice the Lyapunov exponent decreasing to the value of the Lorenz analogue system when adding its data points.
The first time the Lyapunov exponent of the mixed system falls below the three-sigma threshold of the Lyapunov exponent of the training Lorenz data is at marker 2 in Fig. \ref{fig:lorenz-lorenzanalogue}\textbf{c} after roughly $1\,700$ anomalous data points.

The Lorenz analogue system has the same correlation dimension as the Lorenz system\cite{Prosperino2025}.
However, interestingly, mixing the systems results in an increased correlation dimension, with the peak being at marker 3 in Fig. \ref{fig:lorenz-lorenzanalogue}\textbf{c} when the window contains roughly half the Lorenz system and half the Lorenz analogue system.
This bump arises because the two attractors occupy slightly displaced manifolds in phase space, so a mixed window samples pairs of points across the gap between them, and these inter-manifold distances are absorbed by fit for the correlation dimension as an inflated slope, with the effect peaking at equal mixing.
The first time the correlation dimension exceeds the three-sigma interval is at marker 1 after roughly $800$ data points.
Usually, the Lyapunov exponent is the more sensitive measure when classifying chaotic systems, however, in this case, it seems that the correlation dimension is the more sensitive one.
While we are able to detect the introduction of new dynamics using classical, nonlinear measures, we cannot match the high sensitivity of our RC detection algorithm.

Lastly, we benchmark our RC detection algorithm against the three anomaly-detection models introduced earlier, whose configurations we now specify.
The hyperparameters of the benchmark models are chosen to be close to their respective default parameters, while also ensuring that the total number of trainable parameters is roughly comparable to the size of the training dataset.
We did not perform extensive hyperparameter fine-tuning for the benchmark models.
However, we validated that the final configurations were valid by introducing clearly erroneous data in the form of global outliers, to which all models consistently responded by triggering an anomaly.
The failure on the Lorenz-to-Lorenz-analogue switch is therefore qualitative rather than a matter of threshold tuning.
For the GDN model we use a sliding window of $500$ and a single output layer with $256$ nodes.
For the USAD model we use a window size of $50$ and a hidden size of $10$.
The GDN and USAD models are trained on all $10\,000$ normalized data points.
For the ARIMA model, we use a modern online implementation by Kozitsin \textit{et al.}\cite{Kozitsin2021} rather than the classical formulation considered in the review.
It is trained on only $500$ data points, as we found its performance to deteriorate using $5\,000$ training points. 

In Fig. \ref{fig:lorenz-lorenzanalogue}\textbf{d} we present the benchmark results: The deep benchmark models (GDN \& USAD) are not able to detect an anomaly.
Only the shallow ARIMA model is able to detect the anomaly after including a single anomalous data point.
However, after $102$ data points the ARIMA model does not detect the erroneous data anymore, despite its online learning setup.

Overall, among the considered benchmark models, only the RC-based approach reliably and persistently detects the structural switch between the two similar dynamical systems.
The classical nonlinear measures eventually pick up the regime change as well, but only after hundreds to thousands of anomalous data points have accumulated, far beyond the near-instantaneous response of our RC detector.
While the ARIMA model shows an initially fast response and has the lowest computational cost, its detection is not stable.
In contrast, the deep-learning baselines fail to identify the change altogether, despite substantially higher training costs and GPU usage.
For this specific use case, the RC model provides the most favorable trade-off between robustness and computational efficiency.

\subsubsection*{Dependence on hyperparameters}
In this section we analyze the sensitivity of our results to the choice of hyperparameters.
We vary three quantities of interest:
First, the reservoir size $d$, which controls the number of readout weights that form the fingerprint.
Second, the regularization parameter $\beta$, which penalizes large weights and thereby controls their spread.
Lastly, the training length $l$, which determines how much of the attractor is sampled when constructing the fingerprint.

We quantify the training length not in time steps but through the natural-measure coverage $C_\mu$, as it reflects more directly what the reservoir has been exposed to during training.
The natural measure $\mu$ of a chaotic attractor\cite{Eckmann1985} describes the long-time visitation frequency of each region of phase space:
Regions in which the trajectory spends more time receive larger mass.
We discretize the phase space into cubic cells of side $\varepsilon$ and estimate $\mu_i$ from a long reference trajectory as the empirical occupation probability of cell $i$.
The coverage is then defined as the total measure of the cells visited by the training data,
\begin{equation}
\label{equ:natural-measure-coverage}
    C_\mu = \sum_{i \in \mathscr{T}} \mu_i \in [0, 1]\;\;\;,
\end{equation}
where $\mathscr{T}$ is the set of cells touched by at least one training point.
$C_\mu$ captures how much of the dynamically relevant part of the attractor the collection has seen, weighted by how often the system actually visits each region, rather than purely geometric extent.
For all reported experiments we use $\varepsilon = 1.0$.

\begin{figure}[ht]
\centering
\includegraphics{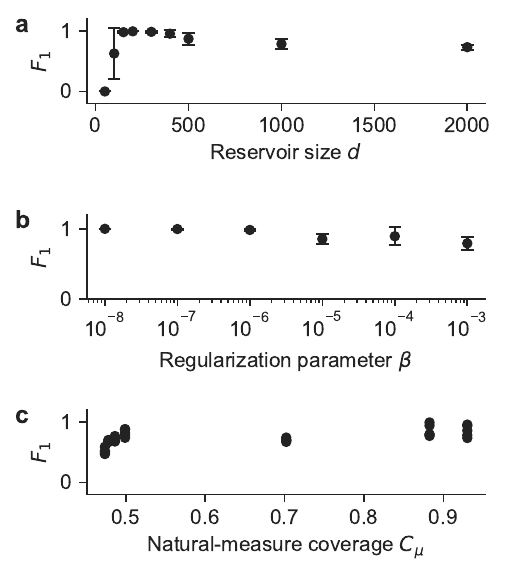}
\caption{\textbf{Dependence on exact hyperparameters.}
Each abscissa shows one analyzed hyperparameter: \textbf{a} the number of nodes of the reservoir $d$, \textbf{b} for the regularization parameter used in the optimization, and \textbf{c} the length of the training data represented by its natural-measure coverage $C_\mu$.
For subfigures \textbf{a} \& \textbf{b} we show the average $F_1$ of 5 runs per hyperparameter.
For \textbf{c} we show each individual run, since $C_\mu$ varies slightly between repeats and does not define discrete bins to average over.
We find our algorithm to be stable against a wide variety of hyperparameters.}%350 words max
\label{fig:hyperparameter-validation}
\end{figure}

We present our results in Fig. \ref{fig:hyperparameter-validation}, where we see that the detector is stable across a wide range of hyperparameters.
In Fig. \ref{fig:hyperparameter-validation}\textbf{a} we analyze the dependence on the reservoir size $d$.
For $d$ in the usual range, the $F_1$ score remains close to $1$ indicating that the fingerprint is well resolved across this range.
Performance at low $d$ is governed by whether the concrete reservoir realization is expressive enough to model the Lorenz system at all.
When it is, detection succeeds reliably.
When it is not, it fails outright. 
This binary behavior across realizations explains the large error bars at small reservoir sizes.

In Fig. \ref{fig:hyperparameter-validation}\textbf{b} we study the dependence on the regularization strength $\beta$ and observe strong performance over a wide range of sensible values, spanning four orders of magnitude from $\beta = 10^{-8}$ to $\beta = 10^{-3}$.
For comparatively large values, $\beta = 10^{-3}$, we note the performance degrading, since strong regularization shrinks all weights toward zero and thereby suppresses the dynamical signal that the fingerprint is meant to encode.

Lastly, Fig. \ref{fig:hyperparameter-validation}\textbf{c} shows the dependence on the natural-measure coverage $C_\mu$ of the training length.
We find the detector to be stable for $C_\mu$ larger than $0.5$, meaning that reliable detection is achieved once the training data has covered the regions accounting for roughly half of the natural measure of the attractor.

Overall, these sweeps confirm that our detector operates reliably over a broad and physically meaningful range of hyperparameters, with failure modes confined to extremes that admit clear mechanistic explanations.

\subsubsection*{Dependence on reservoir computing model}
So far, we follow the classical reservoir computer architecture, which is the most randomness-heavy of the family.
Over the years, more refined reservoir computing architectures have been proposed that reduce or eliminate much of this randomness.
Here, we examine whether our approach of fingerprinting the output matrix generalizes to such architectures, which expose fewer trainable features.
For that, we study the so-called `next generation reservoir computing' architecture (NGRC) by Gauthier \textit{et al.}\cite{Gauthier2021} and our `minimal reservoir computer'\cite{Ma2023} (minRC).
Their implementation details can be found in the Methods.

We apply the same detection algorithm as for the classical RC setup, and use the same amount of training data and the same regularization parameter.
For NGRC we use $k=3$ delay steps spaced at $q=100$ and include nonlinearities up to cubic orders.
For minRC we use a block size of $b=10$, a target spectral radius of $\rho_\textrm{t}=0.1$, and also include nonlinearities up to cubic order in the generalized reservoir states.

\begin{figure}[ht]
\centering
\includegraphics{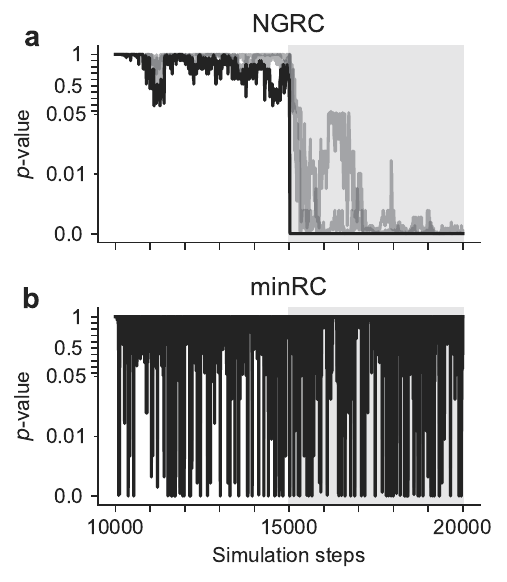}
\caption{\textbf{Performance for different reservoir computing models.}
The upper panel \textbf{a} shows the performance for the `next generation reservoir computer' and the lower panel \textbf{b} shows the performance for the minimal reservoir computer.
In each panel, the bold black line indicates the minimal $p$-value across all coordinates and the gray lines show each individual $p$-value.
We scaled the axes nonlinearly, so the thresholds at $1\,\%$, and $5\,\%$ can be seen better.
Each big axis tick corresponds to an increase of $0.1$.
The grey background again represents the anomalous time series.}%350 words max
\label{fig:hyperparameter-validation-different-rc}
\end{figure}

In Fig. \ref{fig:hyperparameter-validation-different-rc} we present our results, where we immediately note the spiking behavior of minRC in subfigure Fig. \ref{fig:hyperparameter-validation-different-rc}\textbf{b} indicating that minRC is not well suited for anomaly detection.
The minimal design of minRC produces too few readout weights per coordinate, so the resulting ECDFs are coarsely resolved and vary strongly between windows.
This variability widens the reference band $\mathscr{F}$ beyond the point where the Kolmogorov--Smirnov test can separate genuine regime changes from ordinary fingerprint fluctuations, producing the spiking $p$-value seen in Fig. \ref{fig:hyperparameter-validation-different-rc}\textbf{b}.

In contrast, NGRC in subfigure Fig. \ref{fig:hyperparameter-validation-different-rc}\textbf{a} performs well and the combined $p$-value drops to 0 almost immediately after introducing only $18$ anomalous data points.

Our fingerprinting approach therefore generalizes to randomness-reduced architectures as long as the readout exposes enough weights to resolve a stable ECDF, succeeding for `next generation reservoir computer' while failing for the more sparsely parametrized minimal reservoir computer.

\subsection*{Drifting system parameters}
In this section we evaluate the performance of our algorithm in detecting a parametric change in the governing equations.
For that, we use the Lorenz attractor and modify the $\rho$ parameter from its standard value of $\rho=28$ up to $\rho=40$.
We use the same setup as in the previous section.
The training data consists of $10\,000$ data points of the Lorenz system in its standard parametrization.
For the test set, we simulate $5\,000$ steps of the Lorenz system at the standard value of $\rho_\textrm{train}=28$.
After that, we switch to $\rho_\textrm{anomaly}$ and simulate the next $5\,000$ steps.
For our RC detection setup we use the identical setup as described in the previous section.
The same holds for all benchmark models.

\begin{figure}[ht]
\centering
\includegraphics{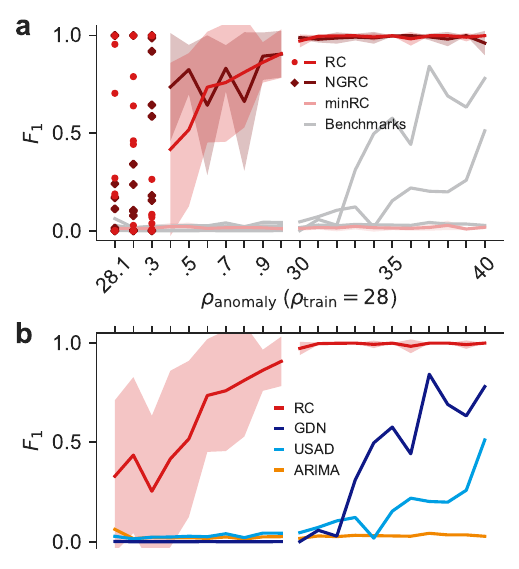}
\caption{\textbf{Detecting a drifting system parameter.}
Along the abscissa of both subfigures the $\rho$ parameter of the anomalous time series is shown.
The ordinate shows the $F_1$ score.
The upper panel $\textbf{a}$ compares different RC architectures with each other, whereas the lower panel $\textbf{b}$ compares the classical RC architecture with the benchmarks.
For the RC-based methods, the shaded band shows the mean with one standard deviation over 5 runs.
The break in each panel marks a change in the scaling of the $\rho$-axis.
To highlight the binary behavior of the $F_1$ score at low $\rho_\textrm{anomaly}$, in panel \textbf{a} we show the individual data points in that region rather than their mean, as the latter would not constitute a meaningful summary.}%350 words max
\label{fig:lorenz-different-rho}
\end{figure}

Comparing the RC architectures in Fig. \ref{fig:lorenz-different-rho}\textbf{a}, we find that the classical RC architecture and NGRC are capable of identifying a transition of the $\rho$ parameter from $\rho=28$ to $\rho_\textrm{anomaly} = 28.1$---albeit very unreliably.
In Fig. \ref{fig:lorenz-different-rho}\textbf{a} for $\rho_\textrm{anomaly} \leq 28.3$, we observe a nearly binary behavior in the individual, non-aggregated $F_1$ scores.
The larger the parameter change, the more reliably both models detect it, and after the anomalous parameter exceeds $\rho_\textrm{anomaly} = 29$, the RC and NGRC detections become virtually perfect.
For a small parameter change the anomalous weight distribution lies only marginally outside the reference band $\mathscr{F}$, so whether the Kolmogorov--Smirnov statistic crosses the detection threshold depends on the particular random reservoir realization.
This effect is visible in Fig. \ref{fig:ecdf_and_d_plot}, where the test ECDF at $\rho_\textrm{anomaly} = 28.1$ barely separates from the reference band $\mathscr{F}$, while larger values of $\rho_\textrm{anomaly}$ produce progressively larger distances $d$ and correspondingly smaller $p$-values.
As $\rho_\textrm{anomaly}$ grows, the fingerprint shifts further beyond the band and every realization detects the change, which is why the $F_1$ score saturates to one.
As before, the minimal reservoir computer is ill-suited to this task.

Among the benchmarks, only GDN is able to detect the anomalies, starting from roughly $\rho=35$, yet its $F_1$ score reaches only $\sim 0.8$ even at $\rho_\textrm{anomaly}=40$ and remains consistently below those of the RC-based detectors, while USAD and ARIMA do not detect the drift anywhere in the swept range.
The RC-based detectors thus resolve parameter drifts roughly seven times smaller than the only competitive benchmark, reliably flagging changes of $\Delta\rho\approx1$ where GDN requires $\Delta\rho\approx7$.

\subsection*{Observational noise}
Our last test on synthetic data consists of introducing noise to the observed data.
For each coordinate we add an independent, additive Gaussian noise with mean $0$ and the standard deviations ranging from $0.01$ to $2$.
Before adding the noise to the time series $\underline{x}{\left(t\right)}$, we calculate the amplitude $A$ of noise and the time series using
\begin{equation}
A = \sqrt{\frac{1}{t} \int_0^t \mathrm{d}t' \left\| \underline{x}{\left( t'\right)} \right\|^2}\;\;\;.
\end{equation}
The signal-to-noise ratio (SNR) is then given by
\begin{equation}
    \textrm{SNR} = \left( \frac{A_\textrm{signal}}{A_\textrm{noise}} \right)^2 = 10\,\log_{10}  \left( \frac{A_\textrm{signal}}{A_\textrm{noise}} \right)^2\,\textrm{dB}\;\;\;.
\end{equation}
The training data consists of $10\,000$ steps of the undisturbed Lorenz system, while the test data consists of $5\,000$ steps of the undisturbed Lorenz followed by $5\,000$ steps of the noisy Lorenz data.
Using the noise mentioned above, we obtain signal strengths ranging from $65\,\textrm{dB}$ down to $18\,\textrm{dB}$.
We leave the setups for all the models unchanged.

\begin{figure}[ht]
\centering
\includegraphics{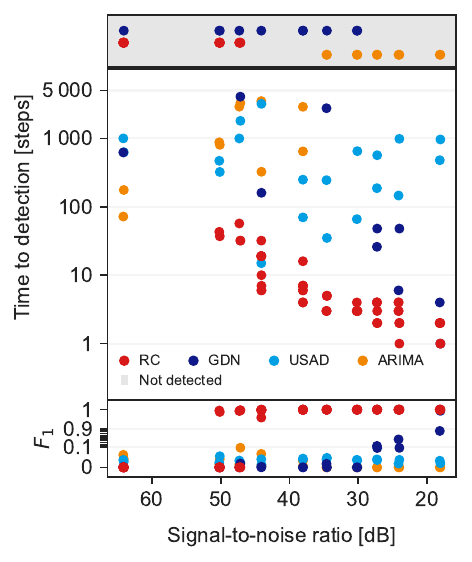}
\caption{\textbf{Detecting barely observable noise.}
The abscissa represents the resulting signal-to-noise ratio from adding the noise on the signal.
The upper panel shows the number of steps after which each algorithm first reported an anomaly being in the anomalous time series.
The lower panel shows the $F_1$ score for each run.
Please note the nonstandard scaling of the axes used for better contextualization of the results.}%350 words max
\label{fig:lorenz-different-noise}
\end{figure}

The results are shown in Fig. \ref{fig:lorenz-different-noise}, where we see the RC model outperforming their benchmarks clearly.
The RC model reliably detects the introduction of noise below a signal strength of $45\,\textrm{dB}$ almost immediately.
This means that the signal is roughly $30\,000$ times stronger than the noise, yet the RC algorithm still detects the anomaly reliably.
Between signal strengths from $45\,\textrm{dB}$ to $50\,\textrm{dB}$---latter means the signal is $100\,000$ times stronger than the noise---we observe a binary behavior of the RC model: Either the detection fails completely or the anomaly is identified almost immediately.
We do not observe a partial detection behavior, where the anomaly is detected late or unreliably.
For a signal strength above $50\,\textrm{dB}$ we are not able to detect the noise.

The benchmark models are not able to detect the noise reliably.
Only for low SNRs, the GDN model seems to perform decently.
The `time to detection' points for benchmark models are more akin to random spikes rather than describing a systematic detection, as underscored by their low $F_1$ score. 

\subsection*{Application to clinical ECG data}
To evaluate our algorithm on real-world data, we require a benchmark suitable to reservoir computing.
Since reservoir computers are particularly effective at modeling complex nonlinear dynamics, electrocardiogram (ECG) recordings are a natural choice, as the underlying cardiac dynamics are known to exhibit nonlinear behavior\cite{Babloyantz1988, Glass2009}.
We sidestep the debate on whether cardiac dynamics are truly chaotic\cite{Glass2009}, as for our use case, nonlinear dynamics are sufficient.
Additionally, we want to state that standard anomaly-detection benchmarks (SWaT\cite{Goh2017}, WaDi\cite{Ahmed2017}, or Yahoo S5 Webscope) are poorly suited to reservoir computing, as their anomalies arise from discrete events or injected outliers rather than deviations from an underlying nonlinear dynamical system.

As our test sample, we use the record 418 from the MIT-BIH Malignant Ventricular Ectopy Database\cite{Greenwald1986} from PhysioNet\cite{Goldberger2000}.
We focus on this record because it contains an episode of ventricular flutter, in which the ventricles depolarize at an abnormally rapid and regular rate.
The resulting ECG loses its characteristic P-QRS-T structure and becomes near-sinusoidal, depicting a qualitative change in the underlying cardiac dynamics---precisely the type of regime shift our algorithm excels at.

\begin{figure}[!ht]
\centering
\includegraphics{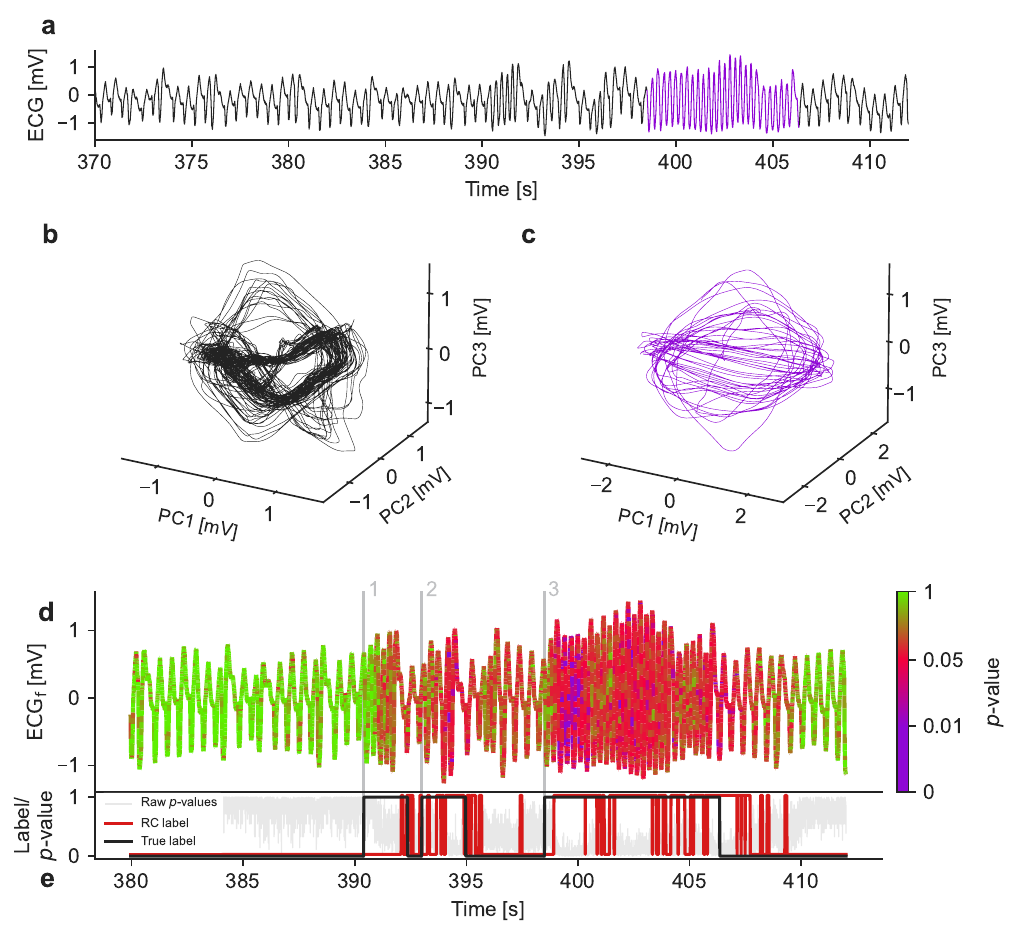}
\caption{\textbf{Applying the reservoir-computer-based detection algorithm to clinical ECG data.}
Subfigure \textbf{a} shows the raw data used in this study.
The purple color highlights a pathological episode of long duration.
Note that not all anomalous heartbeats are marked in purple.
For subfigures \textbf{b} \& \textbf{c} we delay-embedded the filtered data in itself using a time delay of $\tau=84\,\textrm{ms}$ in $8$ dimensions.
We perform a principal component decomposition and show the first three principal components (PC) for the non-pathological data in subfigure \textbf{b} and for the purple anomaly in subfigure \textbf{c}.
Subfigures \textbf{d} \& \textbf{e} show our results:
Subfigure \textbf{d} shows the filtered ECG data with the $p$-value color-coded in each step.
Subfigure \textbf{e} shows the true label in black and the predicted label in red.
In gray we show the raw $p$-value in each step and notice its oscillatory behavior.
} %350 words max
\label{fig:ecg-results}
\end{figure}

The raw ECG signal used in our study is shown in Fig. \ref{fig:ecg-results}\textbf{a}.
Episodes of ventricular flutter can be observed starting at timestep 390 s until 405 s.
A long duration of a ventricular flutter episode is shown in purple.

As standard in ECG signal processing, we apply a fourth-order Butterworth bandpass filter\cite{Butterworth1930} with cutoff frequencies of $0.5\,\mathrm{Hz}$ and $40\,\mathrm{Hz}$ to remove baseline wander, powerline interference, and high-frequency muscle artifacts, while preserving the diagnostically relevant morphology of the signal\cite{Kligfield2007}.
The filtered data is represented as $\textrm{ECG}_\textrm{f}$ in this manuscript.

To provide richer dynamics to the reservoir, we apply Takens' delay embedding theorem\cite{Takens1981}, which guarantees that a dynamical system's attractor can be reconstructed from delayed copies of a single observable.
We embed the ECG time series into a higher-dimensional space using delayed versions of itself, parametrized by a time delay $\tau$ and an embedding dimension $m$. 
The optimal delay $\tau$ is determined by locating the first local minimum of the average mutual information between the signal and its delayed copy\cite{Fraser1986}, which in our case yields $\tau = 84\,\mathrm{ms}$.
The embedding dimension $m$ is selected using the False Nearest Neighbors method\cite{Kennel1992}, choosing the smallest $m$ for which the fraction of false neighbors falls below a predefined threshold, resulting in $m = 8$ for our case.

Figs. \ref{fig:ecg-results}\textbf{b} \& \textbf{c} show the first three principal components of the final $8$-dimensional time series for the non-pathological, baseline data and anomalous data, respectively.
The two phase-space portraits exhibit different geometries: The baseline data traces out a structured, multi-loop trajectory consistent with the regular P-QRS-T cycle, whereas the ventricular flutter regime collapses onto a much simpler, near-closed orbit reflecting the loss of the characteristic ECG morphology. 
This qualitative difference indicates a change in the underlying cardiac dynamics.

Further details on data processing are provided in the `ECG data' section of the Methods.

For our detection method we use a small, Erdős--Rényi-random\cite{Erdos1959} reservoir with $d=100$ nodes and a target spectral radius of $\rho_t = 0.95$.
We feed the whole $8$-dimensional time series into the reservoir, but only regress on the last time step, as described in the `one-dimensional version of reservoir computing' section of the Methods.
For the regression, we apply a noise with a standard deviation of $s=10^{-1}$ on the collected, generalized reservoir states $\mathbf{\tilde{R}}$ (`using noisy reservoir states for real-world data' in the Methods).
We use a regularization parameter of $\beta = 10^{-3}$ and train the reservoir using \textit{only} $1\,000$ time steps.
The collection $\mathscr{F}$ is built simplistically by collecting 2 samples spaced at half the embedding time.
Due to the oscillatory nature of the $p$-value, we define a signal as anomalous if one of the last ten $p$-values has been below $5\,\%$. 
These hyperparameters were not obtained through a systematic grid search or any automated optimization procedure.
Instead, we explored a small number of sensible values by hand and selected a configuration that produced reasonable, yet imperfect, predictive behavior.
We deliberately refrained from extensive tuning, both to keep the setup faithful to the unsupervised, online spirit of the algorithm, and to avoid overfitting our hyperparameter choices to this particular record.

We find our detection algorithm to result in an $F_1$ score of $F_1 = 0.70$.
The detailed results are shown in Figs. \ref{fig:ecg-results}\textbf{d} \& \textbf{e}:
Our algorithm does not produce false positives until marker 1, where a short pathological episode starts.
During the first episode starting at marker 1, we can see the regime of the $p$-value clearly dropping.
However, our detection algorithm is only able to trigger at the end of it, after $1.7\,\textrm{s}$.
Interestingly, our algorithm correctly identifies the very short time between the two episodes as normal heartbeat, and even there the false positives are at a minimum.
The second episode at marker 2 is recognized after $250\,\textrm{ms}$ and its detection sustains the whole duration with the occasional detection oscillation at the beginning.
Between the second and third episodes our algorithm only spikes a single false positive signal, and one can clearly observe the $p$-value rising.
The main episode at marker 3 is detected after $416\,\textrm{ms}$ and detection persists almost the entire duration.
After the main episode, we find the occasional false positive.
We note that the anomaly detection persists consistently even after the pathological episode has ended.
This is explained by the fact that, although the episode itself has stopped, the lookback window still contains anomalous samples and therefore continues to trigger detection.

Finally, we want to emphasize that this experiment is to be understood as a proof-of-concept rather than a competitive ECG-classification exercise.
State-of-the-art supervised approaches on related arrhythmia-detection tasks routinely achieve $F_1$ scores of $0.95$ and higher\cite{Teplitzky2020, Lee2024}.
However, they operate on a fundamentally different framework:
They are trained on large, labeled corpora of annotated pathological episodes (\textit{e.g.}, $783\,\textrm{h}$ in Ref. \citeonline{Teplitzky2020}), they typically perform offline classification of pre-segmented beats or windows, and they are tuned specifically for ECG morphology.
In contrast, our algorithm is fully unsupervised, requires no examples of the pathological regime, is trained on only $8\,\textrm{s}$ of baseline data, runs online, and has not been tuned for ECG signals in any way.
The same machinery used for the synthetic Lorenz experiments is applied here essentially unchanged.

An $F_1$ score of 0.70 obtained under these constraints is therefore not directly comparable to supervised benchmarks.
Instead, it demonstrates that a reservoir-based regime-change detector can identify clinically meaningful transitions in real-world nonlinear dynamics without ever being told what an anomaly looks like.
We see this not as a clinical tool, but rather as a starting point for further work.

\section*{Discussion}
A simple Kolmogorov--Smirnov test on the distribution of reservoir output weights was enough to detect a wide variety of anomalies: from a structural switch between two visually indistinguishable butterfly attractors, through parameter drifts of $\Delta\rho \approx 1$, to additive noise more than four orders of magnitude weaker than the signal, and finally to ventricular flutter in a clinical ECG recording.
Across the synthetic experiments, the RC detector exceeded the sensitivity of deep-learning baselines while training in seconds on a CPU, and on the real-world record it produced an $F_1 = 0.70$ without any tuning to the ECG domain and using only $8\,\textrm{s}$ of baseline data.
Taken together, these results suggest that the readout-weight distribution of a reservoir computer constitutes a compact and informative fingerprint of the underlying dynamics.
This perspective sits in contrast to the dominant use of reservoir computers as forecasting systems.
By studying the trained weights directly rather than the prediction error, we sidestep the question of how far into the future a forecast extends and instead ask whether the same model could have been fitted to the new data at all.

The algorithm is also intrinsically online.
In our framework, each newly arriving sample is simply appended to the most recent training window and triggers a full re-fit of the readout, so streaming detection and training are the same operation rather than two separate modes.
The cost of this design is paid at inference rather than at setup:
Where the deep baselines amortize an expensive training phase against near-instantaneous predictions, our detector pays a small but persistent fitting cost per sample.
For the regimes considered here, sub-second updates are within budget, and the absence of a GPU makes the method well suited to edge deployment and tight energy budgets.

Throughout this work the reference band $\mathscr{F}$ has been treated as static.
We assembled it once from the training window and held it fixed thereafter.
However, nothing in the algorithm requires this: For non-stationary problems, in which the baseline regime itself drifts slowly over time, $\mathscr{F}$ can be updated online by rolling its constituent ECDFs forward as new non-anomalous samples arrive.
This way the reference band tracks the ever evolving baseline while still flagging regime changes.

Although we have framed the algorithm as fully unsupervised, the underlying ECDF over readout weights is also a natural candidate for supervised classification.
Given labeled episodes of distinct dynamical regimes, one could collect a fingerprint per regime and classify a new sample by its Kolmogorov--Smirnov distance to each reference band.
This would convert the present detector into a regime-identification tool and offers a concrete route toward, \textit{e.g.}, distinguishing arrhythmia subtypes in ECG signals rather than merely flagging the onset of a sine rhythm.
We regard this as one of the most immediate directions for future work, alongside a study of how the fingerprint behaves under non-stationary baselines.

Another promising direction concerns the localization of anomalies in coordinate space.
In the noise experiments we were able to flag the perturbation reliably, but not to identify which coordinate had been perturbed, since the classical reservoir computer mixes all inputs into every node through the random matrix $\mathbf{W}_\textrm{in}$.
While NGRC is more structured, it was not expressive enough on a per-coordinate basis to determine the source of the anomaly.
The mere fact that a test as simple as Kolmogorov--Smirnov already yields the detection performance reported here suggests that more sophisticated statistical tests on NGRC's readout should be capable of attributing an anomaly to the specific coordinate that triggered it, turning the present detector into a diagnostic tool.

We want to mention two limitations.
Firstly, the coordinate-wise decomposition of the $p$-value is statistical rather than physical.
Since $\mathbf{W}_\textrm{in}$ randomly mixes all input variables into every reservoir node, a low $p$-value in a given coordinate signals a change in the overall regime, not in a particular coordinate.
Secondly, the detector requires a reservoir that adequately models the system at hand, which can get difficult when including categorical data.

Within these bounds, however, the readout-weight fingerprint appears to capture regime changes that are invisible to both classical nonlinear measures and modern deep-learning detectors, at a fraction of their computational cost.
As the present results illustrate, the trained readout is a rich object in its own right, and treating it as a primary observable---rather than as a means to an end---opens up applications well beyond prediction.

\section*{Methods}
Our detection pipeline combines two ingredients: a reservoir computer, which we use as a compact dynamical model of the input signal, and a statistical test on the distribution of its trained output weights, which serves as the actual anomaly detector.
We describe both components in turn, followed by the data and benchmark methods used in our experiments.

\subsection*{Reservoir computing}
A reservoir computer\cite{Jaeger2001, Maass2002, Jaeger2004} (RC) is a type of recurrent neural network, in which---unlike other recurrent architectures---the recurrent connections are not updated during the training process.
Instead, the input is randomly embedded in a high-dimensional state and then synchronized to the dynamics of the randomly defined reservoir.
The reservoir states encode the dynamics of the input data after synchronization, and are then linearly combined to form the prediction.
It has been discovered that such an architecture is well suited for predicting synthetic chaotic systems\cite{Pathak2017, Lu2018} and also found successful applications in real-world examples\cite{Shahi2021, Brucke2024, Herteux2024}.
We refer to Lukoševičius \& Jaeger\cite{Lukosevicius2009} for a comprehensive review of classical reservoir computing approaches.

Classical reservoir computers map their input $\underline{x} \in \mathbb{R}^n$ randomly onto a $d$-dimensional reservoir $\underline{r}$  using the random mapping $\mathbf{W}_\textrm{in} \in \mathbb{R}^{d \times n}$.
Here, $n$ is the dimensionality of the original system and $d$ is the dimensionality of the reservoir, a hyperparameter of RCs.

The embedded input is then evolved through the reservoir, the core component of an RC, which holds the dynamics to which the embedded input synchronizes.
The reservoir is a network of $d$ randomly connected nodes, whose connectivity is encoded in the adjacency matrix $\mathbf{A} \in \mathbb{R}^{d \times d}$.
Different types of random networks can be used for the reservoir, where typical choices are Barabási--Albert\cite{Barabasi1999} scale-free networks, Erdős--Rényi\cite{Erdos1959} random networks, or Watts--Strogatz\cite{Watts1998} small-world networks.
However, contrary to previous research\cite{Haluszczynski2019}, we find that for our studies that the results are not too dependent on the type of network used for the reservoir, which is why we fix our choice to scale-free networks produced by the Barabási--Albert algorithm with $d = 500$ nodes.
The reservoir is scaled to a target spectral radius of $\rho_{\textrm{t}} = 0.1$.
Unless stated otherwise, this is the standard setup used in this article.

The dynamics of the reservoir is represented by the temporal evolution of the reservoir state $\underline{r}{(t)}\in \mathbb{R}^d$.
The reservoir state is initialized with the zero-vector and iteratively evolved by
\begin{equation}
\label{equ:reservoir-dynamic}
    \underline{r}{(t+1)} = f{\left( \mathbf{A}\,\underline{r}{(t)} + \mathbf{W}_\textrm{in}\,\underline{x}(t)\right)}\;\;\;,
\end{equation}
where $f$ describes a bounded, nonlinear function.
For this article we use the typical choice of the hyperbolic tangent for it.
In order to prevent potential problems stemming from the symmetry of the activation function, we use a quadratic readout as introduced by Herteux \& Räth\cite{Herteux2020}, which means that the squared elements of each reservoir state are appended to form a generalized reservoir state defined by
\begin{equation}
\label{equ:generalized-classicalrc}
    \underline{\tilde{r}}{(t)} =
    \begin{pmatrix}
        \underline{r}{(t)}\\
        \underline{r}{(t)}^2
    \end{pmatrix}\in \mathbb{R}^{2d}\;\;\;,
\end{equation}
where the exponentiation is understood to be applied element-wise.
The output at each point $\underline{y}{(t)}$ is then built as a linear combination of the generalized reservoir state $\underline{\tilde{r}}$ using the linear mapping $\mathbf{W}_{\textrm{out}} \in \mathbb{R}^{n \times 2d}$.

The linear mapping of the generalized reservoir state to the output is the reason why training an RC is a trivial task:
For the training we collect $l$ training samples of the reservoir states and their corresponding output, and store them in the matrices $\tilde{\mathbf{R}} \in \mathbb{R}^{2d \times l}$ and $\mathbf{Y} \in \mathbb{R}^{n \times l}$ respectively.
We then perform a ridge regression to solve the equation $\mathbf{W}_{\textrm{out}}\,\tilde{\mathbf{R}}\,\tilde{\mathbf{R}}^{\textrm{T}} =\mathbf{Y}\,\tilde{\mathbf{R}}^\textrm{T}$ leading to
\begin{equation}
\label{equ:ridge-regression}
    \mathbf{W}_{\textrm{out}} = \mathbf{Y}\,\tilde{\mathbf{R}}^{\textrm{T}}\,\left(\tilde{\mathbf{R}}\,\tilde{\mathbf{R}}^{\textrm{T}} + \beta\,\mathbf{1} \right)^{-1}\;\;\;.
\end{equation}
Here, we use the mathematical trick described by Lukoševičius \& Jaeger\cite{Lukosevicius2009}, where we multiply $\tilde{\mathbf{R}}^{\textrm{T}}$ to the right of $\mathbf{W}_{\textrm{out}}\,\tilde{\mathbf{R}} = \mathbf{Y}$, in order to make the optimization independent of the number of training samples $l$.
In the optimization, $\beta$ is the regularization parameter and $\mathbf{1}$ is the identity matrix.
Unless stated otherwise, in our experiments we use $l = 5\,000$ training samples and a regularization parameter of $\beta = 10^{-6}$.

\subsubsection*{One-dimensional version of reservoir computing}
Reservoir computing performs best on multi-dimensional input data.
For a scalar input time series $x(t)$, we therefore reconstruct the missing dimensions by delay embedding, a principled procedure justified by Takens' theorem\cite{Takens1981}.
Using a time delay $\tau$ and an embedding dimension $m$, the reservoir input at time $t$ becomes the vector of delayed copies
\begin{equation}
    \underline{x}_\textrm{embed.}{(t)} = \left( x(t),\; x(t-\tau),\; \dots,\; x(t-(m-1)\,\tau) \right)^{\textrm{T}} \in \mathbb{R}^m\;\;\;,
\end{equation}
so that the reservoir again receives a multi-dimensional input.

A standard reservoir computer would be trained to forecast this entire vector one step ahead at $\underline{x}_\textrm{embed.}{(t)}$.
However, here this is wasteful:
At time $t+1$, every entry of the target except the first, namely $x(t+1-\tau),\,\dots,\,x(t+1-(m-1)\,\tau)$, has already been observed at time $t$.
Only the unlagged entry $x(t+1)$ carries new information, so we perform the regression on this single target alone.

Concretely, we collect only the unlagged targets in $\underline{Y} \in \mathbb{R}^{1 \times l}$, while still feeding the full embedded input into the reservoir and using the complete generalized reservoir state $\tilde{\mathbf{R}} \in \mathbb{R}^{2d \times l}$.
The ridge regression remains otherwise unchanged,
\begin{equation}
\label{equ:ridge-regression-one-dimensional}
    \underline{W}_{\textrm{out}} = \underline{Y}\,\tilde{\mathbf{R}}^{\textrm{T}}\,\left(\tilde{\mathbf{R}}\,\tilde{\mathbf{R}}^{\textrm{T}} + \beta\,\mathbf{1} \right)^{-1}\;\;\;,
\end{equation}
and yields an output vector $\underline{W}_{\textrm{out}} \in \mathbb{R}^{1 \times 2d}$ that maps all generalized reservoir states onto the single coordinate of interest.

\subsubsection*{Using noisy reservoir states for real-world data}
For real-world recordings, where the signal is itself noisy and the dynamics are only approximately stationary, the linear regression in Eq. \ref{equ:ridge-regression} can become ill-conditioned and produce output weights of large magnitude that are highly sensitive to small changes in the 
training data.
A standard remedy in this situation is to inject a small amount of Gaussian noise during training\cite{Lim2021, Wikner2024}.
This noise injection is known to be equivalent to a form of Tikhonov regularization\cite{Bishop1995}.

We adopt a slight modification of the scheme proposed by Wikner \textit{et al.}\cite{Wikner2024}
Rather than adding noise to the reservoir input at every training step and propagating it through the reservoir dynamics, we add the noise directly to the matrix of generalized reservoir states $\mathbf{\tilde{R}}$ that has already been collected, immediately before solving the ridge regression.

Concretely, before solving for $\mathbf{W}_\textrm{out}$, we replace the matrix of generalized reservoir states $\mathbf{\tilde{R}}$ by a perturbed version
\begin{equation}
    \mathbf{\tilde{R}}_\textrm{noisy} = \mathbf{\tilde{R}} + \boldsymbol{\Xi}\;\;\;.
\end{equation}
Each element of $\boldsymbol{\Xi}$ is drawn independently from a Gaussian distribution with zero mean and standard deviation $s$,
\begin{equation}
    \xi_{ij} \sim \mathscr{N}{\left(0,\, s^2\right)}\;\;\;.
\end{equation}
The noise level $s$ is treated as an additional hyperparameter of the model.

\subsection*{Modern reservoir computing architectures}
The classical reservoir computer is heavily randomized: both the input mapping $\mathbf{W}_\textrm{in}$ and the reservoir adjacency matrix $\mathbf{A}$ are drawn at random, and only the readout is fitted to the data.
Two recent architectures aim to reduce this randomness in complementary ways:
`next generation reservoir computing' (NGRC) by Gauthier \textit{et al.}\cite{Gauthier2021} dispenses with the recurrent reservoir entirely and replaces it by a deterministic feature vector built from time-delayed copies of the input, whereas the minimal reservoir computer (minRC)\cite{Ma2023} preserves the recurrent structure but enforces a sparse, block-structured connectivity that drastically reduces the number of random parameters.
We briefly summarize both architectures below and refer to the original articles for a complete description.

\subsubsection*{Minimal reservoir computing}
The minimal reservoir computer\cite{Ma2023} does not embed the input into a random, high-dimensional space, instead, the input matrix $\mathbf{W}_\textrm{in}$ is block-structured, so that each feature $f$ is fed into its own disjoint block of $b$ reservoir nodes.
The features $f$ consist of the input variables, and linear combinations thereof: \textit{e.g.}, $f \in\left\{ x_1, x_2, x_3, x_1+x_2,\,\dots, x_1+x_2+x_3\right\}$ for a three-dimensional system.
The block $\mathbf{J}_f$ for a feature $f$ then consists of a fully connected network represented by a matrix of $1$ and the reservoir $\mathbf{A}$ then has the form of
\begin{equation}
\mathbf{A} = \frac{\rho_\textrm{t}}{b}\,\diag{\left(\mathbf{J}_{x_1},\,\dots,\,\mathbf{J}_{f},\,\dots,\,\mathbf{J}_{x_1+x_2+x_3}\right)}\;\;\;.
\end{equation}
The reservoir is still scaled to a target spectral radius $\rho_\textrm{t}$.
The dynamical evolution of the reservoir states is still the same as in Eq. \ref{equ:reservoir-dynamic}, with the nonlinear activation function $f$ replaced by the identity function.

To introduce nonlinearity, before training the reservoir states are generalized by appending them with element-wise powers of themselves up to nonlinearity degree, which is cubic in this article.
The readout is trained by ridge regression on the generalized reservoir states following Eq. \ref{equ:ridge-regression}.
Unless stated otherwise, we use a block size of $b = 10$ and a target spectral radius of $\rho_\textrm{t} = 0.1$.

\subsubsection*{`Next generation reservoir computing'}
`Next generation reservoir computing'\cite{Gauthier2021} eliminates the traditional concept of a reservoir altogether.
Instead of synchronizing the input to a network, the feature vector at time $t$ is constructed deterministically from $k$ time-delayed copies of the input $\underline{x}{(t)},\,\underline{x}{(t-q)},\,\dots,\,\underline{x}{(t-(k-1)\,q)}$, spaced by $q$ time steps, together with all unique monomials of these delayed inputs up to a chosen polynomial order.
The resulting feature vector takes the role of the generalized reservoir state $\underline{\tilde{r}}{(t)}$, and the readout $\mathbf{W}_\textrm{out}$ is obtained via the same ridge regression as in Eq. \ref{equ:ridge-regression}.
The only hyperparameters are the number of delays $k$, the spacing $q$, the polynomial order, and the regularization strength $\beta$.
No random matrices are involved.
Unless stated otherwise, we use $k = 3$ delays, a spacing of $q = 100$ time steps, and monomials up to cubic order.

\subsection*{Anomaly detection}
The core idea of our anomaly detection method is to treat the output weights $\mathbf{W}_\textrm{out}$ of a trained reservoir computer as a signature of the underlying dynamics: Since the readout is the only component that adapts to the data, its weights have to encode the dynamical regime to which the reservoir has been synchronized.
We summarize this signature through the empirical cumulative distribution function (ECDF) of the entries of $\mathbf{W}_\textrm{out}$, which serves as a `fingerprint' of the trained system.
During an initial reference phase, we apply a sliding window to the training data and retrain the reservoir on each window, yielding a collection of reference ECDFs $\mathscr{F} = \left\{F_1, \dots, F_t\right\}$ that describes the natural variability of the fingerprint under normal dynamics.
In the subsequent online phase, each newly arriving sample is appended to the most recent training window and the reservoir is retrained, producing an updated ECDF $E$.
We then measure the discrepancy between $E$ and $\mathscr{F}$ using a modified Kolmogorov--Smirnov test\cite{Kolmogorov1933, Smirnov1948}, which assigns a $p$-value to each new sample.
We compute this $p$-value separately for each row of $\mathbf{W}_\textrm{out}$, \textit{i.e.} one ECDF and one $p$-value per output coordinate, and define the overall $p$-value of the sample as the minimum across all coordinates.
We emphasize that this per-row decomposition is statistical only and does not admit a coordinate-wise interpretation of the anomaly.
Since the input matrix $\mathbf{W}_\textrm{in}$ randomly mixes all input variables into every reservoir node, the weights in a given row of $\mathbf{W}_\textrm{out}$ do not isolate the dynamics of any single input coordinate.
A low $p$-value in one row therefore signals that the overall regime has changed, not which input variable drove the change.
A small $p$-value indicates that the current weight distribution is incompatible with the reference and is therefore flagged as anomalous.

\begin{figure}[!ht]
\centering
\includegraphics{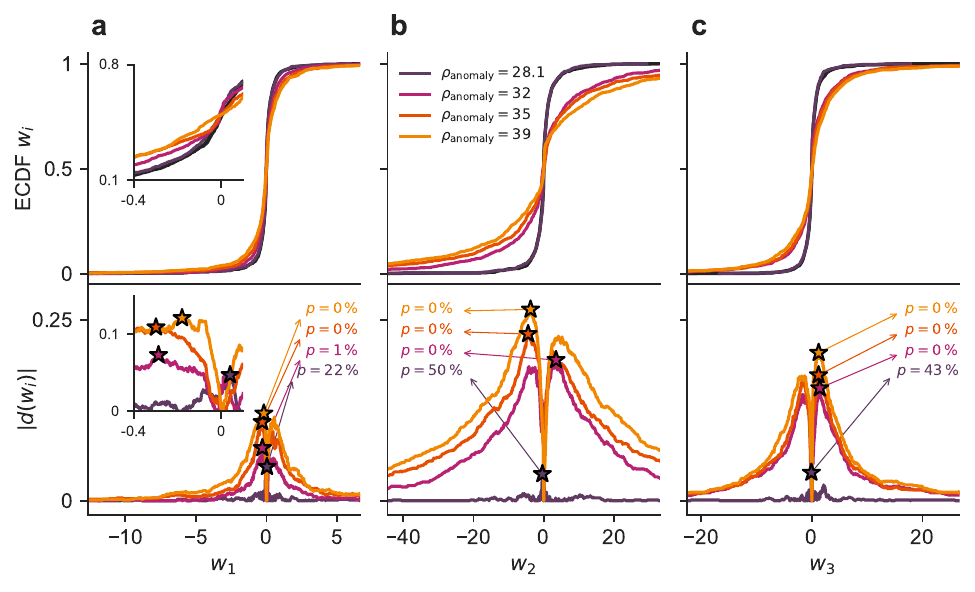}
\caption{\textbf{ECDFs as fingerprints with corresponding distance $d$ and $p$-value calculation for drifting Lorenz parameters.}
Each column corresponds to one coordinate of the trained readout matrix $\mathbf{W}_\textrm{out}$.
Subfigures \textbf{a}, \textbf{b}, and \textbf{c} show the readout weights $w_1$, $w_2$, and $w_3$ corresponding to the coordinates $x_1$, $x_2$, and $x_3$, respectively.
The upper panel of each subfigure shows the empirical cumulative distribution function (ECDF) of the corresponding row of $\mathbf{W}_\textrm{out}$.
The lower panel shows the distance $\left| d(w_i) \right|$ between the test ECDF and the reference band spanned by $F_\textrm{min}$ and $F_\textrm{max}$.
The colored lines correspond to four anomalous configurations with $\rho_\textrm{anomaly} \in \{28.1, 32, 35, 39\}$.
The legend in subfigure \textbf{b} applies to all panels.
The stars in the lower panels mark the supremum of $\left| d(w_i) \right|$, which defines the Kolmogorov--Smirnov statistic for that coordinate, and the colored labels report the associated $p$-values.
} %350 words max %350 words max
\label{fig:ecdf_and_d_plot}
\end{figure}

Fig. \ref{fig:ecdf_and_d_plot} illustrates this procedure for the Lorenz system under a drifting $\rho$ parameter.
The top row shows the reference band of ECDFs $\mathscr{F}$ with four test ECDFs $E$ obtained at increasing $\rho_\textrm{anomaly}$, and the bottom row shows the associated distance $\left| d(w_i) \right|$ per coordinate, whose supremum---marked by the stars---defines the Kolmogorov--Smirnov statistic and thus the reported $p$-values.

\subsubsection*{Kolmogorov--Smirnov test}
In order to quantify whether a drawn ECDF stems from a reference CDF, we use the Kolmogorov--Smirnov test.
Given a sample ECDF $E{(w)}$ and a reference CDF $F{(w)}$, the test proposes a null hypothesis of the sample $E$ stemming from $F$ and thus, the distributions being the same.

In order to test the null hypothesis, we need the Kolmogorov--Smirnov statistic $D$, which is defined by
\begin{equation}
    D = \sup_w \left| E{(w)} - F{(w)}\right|\;\;\;.
\end{equation}
The statistic describes the largest absolute difference between the two CDFs.

Under the assumption of the null hypothesis, the scaled statistic $D_n$ will follow the Kolmogorov distribution $K$ as the sample size $n$ goes to infinity.
The statistic $D_n$ is scaled by
\begin{equation}
    D_n = \sqrt{n}\,D\;\;\;.
\end{equation}
The CDF of the Kolmogorov distribution is given by
\begin{equation}
    \mathscr{P}{(K \leq w)} = 1 - 2\,\sum_{k=1}^\infty(-1)^{k-1}\,\exp{\left( -2\,k^2\,w^2\right)}\;\;\;.
\end{equation}
The $p$-value of the test is given by
\begin{equation}
\label{equ:p-value}
    p = 1 - \mathscr{P}{\left(K \leq D_n \right)}\;\;\;.
\end{equation}
The $p$-value gives the probability of obtaining a test statistic at least as extreme as $D_n$ under the null hypothesis that $E$ is drawn from $F$.
A small $p$-value therefore allows the null hypothesis to be rejected at the chosen significance level.
In simpler terms, a small $p$-value indicates that the observed sample is unlikely to have been produced by the reference distribution $F$, and that $E$ and $F$ should be considered different.

The convergence of $D_n$ is relatively slow, which is why a large sample size $n$ is required\cite{Massey1951,Marsaglia2003}.
However, for typical parametrizations of RCs we obtain weights in the number of hundreds for each variable.

Vrbik\cite{Vrbik2018} discovered a correction for this purpose, which allows for a useful estimation of the statistic under a small sample size.
He proposed replacing $w$ with the correction function $c{(w)}$.
The function is given by
\begin{equation}
    c{(w)} = w + \dfrac{1}{6\,\sqrt{n}} + \dfrac{w-1}{4\,n}
\end{equation}
with $n$ being the sample size.
This correction changes Eq. \ref{equ:p-value}, the definition of the $p$-value, to
\begin{equation}
    p = 1 - \mathscr{P}{\left(K \leq c{\left(D_n\right)} \right)}\;\;\;.
\end{equation}
We apply this correction if the number of weights per coordinate is below $1\,000$.

\subsubsection*{Extension to uncertain reference distribution}
The Kolmogorov--Smirnov test assumes perfect knowledge of the reference CDF $F$.
However, in our framework we do not have a single reference CDF.
Instead, due to the rolling window approach, we have a collection of CDFs $\mathscr{F} = \left\{ F_{1}, ..., F_{t} \right\}$.
For this reason, we need to modify the statistic $D$.
Since each $F$ represents a valid distribution of weights, we define a band for each point on the domain stemming from the minimal value to the maximal value of the observed collection $\mathscr{F}$.
Let $\supp$ be the support of a function, we can define the effective global domain $w_\textrm{G}$ as
\begin{equation}
    w_\textrm{G} = \bigcup_{F \in \mathscr{F}}\supp{(F)}\;\;\;.
\end{equation}
We then define the lower band $F_\textrm{min}$ as the minimal value across the effective global domain, and the upper band $F_\textrm{max}$ as the maximal value accordingly:
\begin{subequations}
\label{equ:fmin-fmax-calculation}
\begin{align}
&F_\textrm{min} = \left\{ \min_{F\in \mathscr{F}} F{(w)} \mid w \in w_\textrm{G} \right\}\\
&F_\textrm{max} = \left\{ \max_{F\in \mathscr{F}} F{(w)} \mid w \in w_\textrm{G} \right\}\;\;\;.
\end{align}
\end{subequations}
We are now able to define a distance $d{(w)}$ between an empirical CDF $E$ and a collection of reference CDFs $\mathscr{F}$ as
\begin{equation}
    d{(w)} =
    \begin{cases}
    E{(w)} - F_\textrm{min}{(w)} & \text{if}\;\;E{(w)} < F_\textrm{min}{(w)}\\
    0              & \textrm{if}\;\;F_\textrm{min}{(w)} \leq E{(w)} \leq F_\textrm{max}{(w)}\\
    E{(w)} - F_\textrm{max}{(w)}& \text{if}\;\;E{(w)} > F_\textrm{max}{(w)}\;\;\;.
\end{cases}
\end{equation}
The statistic $D$ then becomes
\begin{equation}
    D = \sup_w \left| d(w) \right|\;\;\;.
\end{equation}
With this modification, the empirical CDF $E$ is only penalized when it falls outside the band spanned by the reference collection $\mathscr{F}$, so that the natural variability across rolling-window references is absorbed into the normal regime rather than flagged as anomalous.
The modified statistic $D$ is then fed into the Kolmogorov--Smirnov framework to obtain a $p$-value for each newly arriving sample.

\subsection*{Data}
We evaluate our algorithm on both synthetic and real-world nonlinear data.
The synthetic experiments use the Lorenz system\cite{Lorenz1963}, the canonical choice for nonlinear dynamics, together with a second system designed to produce a visually indistinguishable attractor.
The real-world test uses clinical electrocardiogram (ECG) recordings, described in the corresponding subsection below.

All synthetic data are integrated using the explicit Runge--Kutta method of order 5(4)\cite{Dormand1980} with a step size of $\Delta t = 0.01$.
Initial conditions are drawn uniformly from the integers between $-20$ and $20$ for each coordinate, and the first $10\,000$ steps are discarded as transient.

\subsubsection*{Lorenz system}
Initially introduced as a model for atmospheric convection, the Lorenz system\cite{Lorenz1963} is described by
\begin{subequations}
\label{equ:lorenz-equation}
\begin{align}
\dot{x}_1&=-\sigma\,x_1 + \sigma\,x_2\\
\dot{x}_2&=\rho\,x_1 - x_2 - x_1\,x_3\\ 
\dot{x}_3&=-\beta\,x_3 + x_1\,x_2\;\;\;.
\end{align}
\end{subequations}
We use the standard parametrization $\sigma=10$, $\rho=28$, and $\beta=\tfrac{8}{3}$.
For analyzing the capability of detecting changing parameters, we vary $\rho$, which is known to control the shape of the attractor.

\subsubsection*{Alternative realization of Lorenz attractor: Lorenz analogue system}
In Ref. \citeonline{Prosperino2025}, we tried to reconstruct the governing equations of a system using a skeleton system containing a defined library of terms, in our example all polynomial terms up to second order (Eqs. \ref{equ:library-order-2}), and a measured trajectory.
Our approach was to synchronize the skeleton system to the data.
Once synchronized, we used gradient descent algorithms to change the parameters $\theta_{i\,j}$ to the direction minimizing the distance between the simulated trajectory and the original trajectory.
While we were not able to reconstruct the original Lorenz equations, we found different sets of parameters $\boldsymbol{\Theta}$ which describe the Lorenz system well and produce the same-looking butterfly-shaped attractor, as can be seen in the projection in Fig. \ref{fig:lorenz-lorenzanalogue}\textbf{a}.
Eq. \ref{equ:parameters-library} shows the one set of parameters used in this study.

\begin{subequations}
\label{equ:library-order-2}
\begin{align}
\dot{x}_1&=\theta_{1\,1} + \theta_{1\,2}\,x_1 + \theta_{1\,3}\,x_2 + \theta_{1\,4}\,x_3 + \theta_{1\,5}\,x_1^2 + \theta_{1\,6}\,x_1\,x_2 + \theta_{1\,7}\,x_1\,x_3 + \theta_{1\,8}\,x_2^2 + \theta_{1\,9}\,x_2\,x_3 + \theta_{1\,10}\,x_3^2\\
\dot{x}_2&=\theta_{2\,1} + \theta_{2\,2}\,x_1 + \theta_{2\,3}\,x_2 + \theta_{2\,4}\,x_3 + \theta_{2\,5}\,x_1^2 + \theta_{2\,6}\,x_1\,x_2 + \theta_{2\,7}\,x_1\,x_3 + \theta_{2\,8}\,x_2^2 + \theta_{2\,9}\,x_2\,x_3 + \theta_{2\,10}\,x_3^2\\ 
\dot{x}_3&=\theta_{3\,1} + \theta_{3\,2}\,x_1 + \theta_{3\,3}\,x_2 + \theta_{3\,4}\,x_3 + \theta_{3\,5}\,x_1^2 + \theta_{3\,6}\,x_1\,x_2 + \theta_{3\,7}\,x_1\,x_3 + \theta_{3\,8}\,x_2^2 + \theta_{3\,9}\,x_2\,x_3 + \theta_{3\,10}\,x_3^2
\end{align}
\end{subequations}

\begin{equation}
\label{equ:parameters-library}
\boldsymbol{\Theta} =
\begin{pmatrix}
-12.74 & -0.6302 & 4.258 & 1.905 & 0.3727 & -0.3356 & -0.2582 & 0.07002 & 0.1586 & -0.06479\\
-14.66 & 12.73 & 8.437 & 1.924 & 0.1941 & -0.1544 & -0.5768 & 0.01826 & -0.2674 & -0.05295\\
-19.67 & 1.037 & -0.6941 & -0.02220 & 0.3800 & 0.6524 & -0.02853 & 0.07524 & 0.02042 & -0.08016
\end{pmatrix}
\end{equation}

\subsubsection*{ECG data}
As a real-world application, we test our algorithm on clinical ECG data.
We use the record 418 from the MIT-BIH Malignant Ventricular Ectopy Database\cite{Greenwald1986} from PhysioNet\cite{Goldberger2000}, taking the first of its two ECG leads in physical units, millivolts.

We then apply a fourth-order Butterworth bandpass filter\cite{Butterworth1930} to remove baseline wander, powerline interference, and high-frequency muscle artifacts.
Specifically, we apply the Fourier transformation $\textrm{FT}$ to the original signal $\textrm{ECG}{(t)}$ to obtain $\hat{\textrm{ECG}}{(f)} = \textrm{FT}{\left[ \textrm{ECG}{(t)} \right]}$.
Each mode is then multiplied by $\left| H{(f)} \right|^2$ with
\begin{equation}
\left| H{(f)} \right|^2 = \dfrac{1}{1 + \left( \nicefrac{f_\textrm{low}}{f}\right)^{2\cdot4}} \; \dfrac{1}{1 + \left( \nicefrac{f}{f_\textrm{high}} \right)^{2\cdot4}}\;\;\;,
\end{equation}
with $f_\textrm{low} = 0.5\,\textrm{Hz}$ and $f_\textrm{high} = 40\,\textrm{Hz}$.
Finally, we transform the result back to the time domain to obtain the filtered signal $\textrm{ECG}_\textrm{f}$ using
\begin{equation}    
\textrm{ECG}_\textrm{f}{(t)} = \textrm{FT}^{-1}\left[\left| H{(f)} \right|^2\,\,\hat{\textrm{ECG}}{(f)}\right].
\end{equation}

To provide the reservoir with a richer, multi-dimensional input than a scalar time series, we apply Takens' delay-embedding theorem\cite{Takens1981}, which guarantees that, for almost every choice of observable, time delay $\tau$, and sufficiently large embedding dimension $m$ the trajectory
\begin{equation}
    \underline{x}{(t)} = \left(\begin{array}{l}
        \textrm{ECG}_\textrm{f}(t)\\
        \textrm{ECG}_\textrm{f}(t - \tau)\\
        \quad\quad\vdots\\
        \textrm{ECG}_\textrm{f}(t - (m-1)\,\tau)
    \end{array}\right) \in \mathbb{R}^m\;\;\;.
\end{equation}
is diffeomorphic to the attractor of the underlying dynamical system.
However, we still regress to $\underline{y}{(t+1)} = \textrm{ECG}_\textrm{f}(t+1) \in \mathbb{R}$.

The time delay $\tau$ and embedding dimension $m$ are determined directly using the filtered ECG data.
The time delay is selected by the criterion of Fraser \& Swinney\cite{Fraser1986}, namely as the first local minimum of the average mutual information $I{(\tau)}$ between $\textrm{ECG}_\textrm{f}{(t)}$ and $\textrm{ECG}_\textrm{f}{(t-\tau)}$.
Writing $a$ and $a'$ for the signal values at times $t$ and $t-\tau$, the average mutual information is defined by
\begin{equation}
\label{equ:ami}
    I{(\tau)} = \iint \textrm{d}a\,\textrm{d}a'\,p{(a, a'; \tau)}\,\log{\dfrac{p{(a, a'; \tau)}}{p{(a)}\,p{(a')}}}\;\;\;,
\end{equation}
where $p{(a)}$ and $p{(a')}$ are the marginal probability densities of $\textrm{ECG}_\textrm{f}{(t)}$ and $\textrm{ECG}_\textrm{f}{(t-\tau)}$, respectively, and $p{(a, a'; \tau)}$ is their joint density at lag $\tau$.
For our record, this yields $\tau = 84\,\textrm{ms}$.
The embedding dimension is then selected via the False Nearest Neighbors algorithm of Kennel \textit{et al.}\cite{Kennel1992}:
For each candidate $m$, we identify in the embedded space the nearest neighbor of every point and check whether adding the $(m+1)$-th coordinate causes that neighbor to move away by more than a factor $R_\textrm{tol} = 15$ of its current distance, or to leave a global ball of radius $A_\textrm{tol} = 2$ standard deviations.
Neighbors satisfying either criterion are flagged as false, indicating that they are close in $\mathbb{R}^m$ only because the embedding dimension is too low to resolve the true geometry of the attractor.
We choose the smallest $m$ for which the false-neighbor fraction drops below $1\,\%$, obtaining $m = 8$.

As training data, we use $1\,000$ time steps preceding the first ventricular flutter episode ranging from $376\,\textrm{s}$ to $384\,\textrm{s}$.

\subsection*{Benchmark methods}
We compare our algorithm against three benchmark methods, selected based on the recent review by Mejri \textit{et al.}\cite{Mejri2024} of unsupervised anomaly detection in multivariate time series.
The review groups anomalies into several types and reports, for each method, those it handles best.
Since the regime changes considered in this study---a switch between dynamical systems, a drifting parameter, and the addition of observational noise---are most naturally described as shapelet or seasonality anomalies, we restrict our choice to methods that perform well on these categories.
This leaves two predictive approaches, ARIMA and the Graph Deviation Network (GDN)\cite{Deng2021}, which the review highlights as having the best generalization across anomaly types, and the reconstruction-based USAD\cite{Audibert2020}, which is reported to be particularly robust to shapelet and seasonality anomalies.
Together, these three methods span the predictive and reconstruction-based families of unsupervised detectors and provide a representative basis for comparison.
In the following, we briefly summarize each method and specify our configuration.
We refer to the original publications for a complete description.

\subsubsection*{GDN (Graph Deviation Network)}
The Graph Deviation Network\cite{Deng2021} is a predictive, graph-based approach to multivariate anomaly detection.
Each variable of the input time series is represented as a node, and a sparse graph encoding the pairwise dependencies between variables is learned jointly with the model from the training data.
At inference time, the value of each variable is predicted from the recent history of its graph neighbors using a graph attention mechanism, and the deviation between the predicted and observed values defines a per-node anomaly score.
The overall anomaly score at each time step is taken as the maximum of the per-node scores, so that an anomaly localized in a single variable is sufficient to flag the entire sample.
We use the reference implementation provided by the authors\cite{Deng2021} with a sliding window of $500$ time steps and a single output layer of $256$ units, and train the model on all $10\,000$ normalized training points.
The remaining hyperparameters are kept at their default values.

\subsubsection*{USAD (UnSupervised Anomaly Detection)}
USAD\cite{Audibert2020} is a reconstruction-based approach that combines an autoencoder architecture with an adversarial training scheme inspired by generative adversarial networks.
The model consists of one shared encoder and two decoders, which are trained jointly in two phases:
First, both autoencoders are trained to reconstruct the input.
Second, they are trained adversarially, so that one decoder learns to reconstruct the input while the other learns to distinguish real inputs from those reconstructed by the first.
At inference time, the anomaly score of a sample is defined as a weighted combination of the reconstruction errors of both autoencoders, with the adversarial component amplifying the score on inputs that deviate from the training distribution.
We use the reference implementation provided by the authors\cite{Audibert2020} with a sliding window of $50$ time steps and a hidden size of $10$, and train the model on all $10\,000$ normalized training points.
The remaining hyperparameters are kept at their default values.

\subsubsection*{ARIMA}
The AutoRegressive Integrated Moving Average (ARIMA) model is a classical, predictive approach to time series modeling, in which each value of a time series is expressed as a linear combination of its past values and past forecast errors.
While ARIMA has long been used for anomaly detection by thresholding the residual between the model's forecast and the observed value, we adopt the more recent variant of Kozitsin \textit{et al.}\cite{Kozitsin2021}, in which the moving-average terms are absorbed into additional autoregressive terms and the resulting weight vector is updated online via a gradient-descent rule.
In this formulation, the discrete differences of the weight vector along the time axis vanish in the stationary regime and deviate from zero whenever the dynamics change, so that the anomaly score is derived directly from the evolution of the model parameters rather than from prediction residuals.
We use the reference implementation provided by the authors\cite{Kozitsin2021} and train the model on $500$ time steps, as we found its detection performance to deteriorate when training on the full training set.
The remaining hyperparameters are kept at their default values.

\bibliography{bibliography}

@article{Prosperino2025,
    year = {2025},
    volume = {6},
    pages = {015012},
    author = {Davide Prosperino and Haochun Ma and Christoph Räth},
    title = {A generalized method for estimating parameters of chaotic systems using synchronization with modern optimizers},
    journal = {J. Phys. Complex.}}

@article{Lorenz1963,
    title = {Deterministic {N}onperiodic {F}low},
    journal = {JAS},
    volume = {20},
    number = {2},
    pages = {130-141},
    year = {1963},
    author = {Edward N. Lorenz}}

@inproceedings{Deng2021,
  title = {Graph {N}eural {N}etwork-{B}ased {A}nomaly {D}etection in {M}ultivariate {T}ime {S}eries},
  author = {Ailin Deng and Bryan Hooi},
  booktitle = {Proceedings of the Thirty-Fifth AAAI Conference on Artificial Intelligence (AAAI-21)},
  year = {2021},
  pages = {4027-4035}
}

@inproceedings{Lai2021,
  title = {Revisiting {T}ime {S}eries {O}utlier {D}etection: {D}efinitions and {B}enchmarks},
  author = {Kwei-Herng Lai and Daochen Zha and Junjie Xu and Yue Zhao and Guanchu Wang and Xia Ben Hu},
  booktitle = {Proceedings of the Neural Information Processing Systems Track on Datasets and Benchmarks},
  year = {2021}
}

@article{Mejri2024,
    title = {Unsupervised anomaly detection in time-series: {A}n extensive evaluation and analysis of state-of-the-art methods},
    journal = {Expert Syst. Appl.},
    volume = {256},
    pages = {124922},
    year = {2024},
    author = {Nesryne Mejri and Laura Lopez-Fuentes and Kankana Roy and Pavel Chernakov and Enjie Ghorbel and Djamila Aouada}}

@article{Kozitsin2021,
    title = {Online {F}orecasting and {A}nomaly {D}etection {B}ased on the {ARIMA} {M}odel},
    journal = {Appl. Sci.},
    volume = {11},
    pages = {3194},
    year = {2021},
    author = {Viacheslav Kozitsin and Iurii Katser and Dmitry Lakontsev}}

@inproceedings{Audibert2020,
  title = {{USAD}: {UnSuper}vised {A}nomaly {D}etection on {M}ultivariate {T}ime {S}eries},
  author = {Julien Audibert and Pietro Michiardi and Frédéric Guyard and Sébastien Marti and Maria A. Zuluaga},
  booktitle = {Proceedings of the 26th ACM SIGKDD Conference on Knowledge Discovery and Data Mining (KDD '20)},
  year = {2020},
  pages = {3395-3404}
}

@techreport{Jaeger2001,
    author = {Herbert Jaeger},
    title = {The ``echo state'' approach to analysing and training recurrent neural networks - with an {E}rratum note},
    institution = {GMD Forschungszentrum Informationstechnik},
    year = {2001}
}

@article{Maass2002,
  title={Real-time computing without stable states: a new framework for neural computation based on perturbations},
  author={Wolfgang Maass and Thomas Natschläger and Henry Markram},
  journal={Neural Comput.},
  volume={14},
  number={11},
  pages={2531-2560},
  year={2002},
}

@article{Jaeger2004,
  title = {{H}arnessing {N}onlinearity: {P}redicting {C}haotic {S}ystems and {S}aving {E}nergy in {W}ireless {C}ommunication},
  author={Herbert Jaeger and Harald Haas},
  journal={Science},
  volume={304},
  number={5667},
  pages={78-80},
  year={2004},
}

@article{Lukosevicius2009,
    author = {Mantas Lukoševičius and Herbert Jaeger},
    title = {Reservoir computing approaches to recurrent neural network training},
    journal = {Comput. Sci. Rev.},
    year = {2009},
    pages = {127-149},
    volume = {3}
}

@article{Barabasi1999,
    author = {Albert-László Barabási and Réka Albert},
    title = {{E}mergence of {S}caling in {R}andom {N}etworks},
    journal = {Science},
    year = {1999},
    volume = {286},
    issue = {5439},
    pages = {509-512}
}

@article{Babloyantz1988,
    author = {Agnessa Babloyantz and Alain Destexhe},
    title = {{I}s the {N}ormal {H}eart a {P}eriodic {O}scillator?},
    journal = {Biol. Cybern.},
    year = {1988},
    volume = {58},
    pages = {203-211}
}

@article{Glass2009,
    author = {Leon Glass},
    title = {{I}ntroduction to {C}ontroversial {T}opics in {N}onlinear {S}cience: {I}s the {N}ormal {H}eart {R}ate {C}haotic?},
    journal = {Chaos},
    year = {2009},
    volume = {19},
    pages = {028501}
}

@inproceedings{Goh2017,
  title = {A {D}ataset to {S}upport {R}esearch in the {D}esign of {S}ecure {W}ater {T}reatment {S}ystems},
  author = {Jonathan Goh and Sridhar Adepu and Khurum Nazir Junejo and Aditya Mathur },
  booktitle = {Critical Information Infrastructures Security (CRITIS 2016)},
  year = {2017},
  pages = {88-99}
}

@inproceedings{Ahmed2017,
  title = {{WADI}: {A} {W}ater {D}istribution {T}estbed for {R}esearch in the {D}esign
of {S}ecure {C}yber {P}hysical {S}ystems},
  author = {Chuadhry Mujeeb Ahmed and Venkata Reddy Palleti and Aditya P. Mathur},
  booktitle = {3rd International Workshop on Cyper-Physical Systems for Smart Water Networks (CySWATER '17)},
  year = {2017},
  pages = {25-28}
}

@thesis{Greenwald1986,
  author = {S. D. Greenwald},
  title = {Development and analysis of a ventricular fibrillation detector},
  school = {Massachusetts Institute of Technology},
  year = {1986},
  type = {M.S. Thesis}
}

@article{Goldberger2000,
    author = {A. Goldberger and L. Amaral and L. Glass and J. Hausdorff and P. C. Ivanov and R. Mark and J.E. Mietus and G. B. Moody and C. K. Peng and H.E. Stanley},
    title = {{PhysioBank}, {PhysioToolkit}, and {PhysioNet}: {C}omponents of a new research resource for complex physiologic signals},
    journal = {Circulation},
    year = {2000},
    volume = {101},
    pages = {e215-e220}
}

@article{Fraser1986,
    author = {Andrew M. Fraser and Harry L. Swinney},
    title = {Independent coordinates for strange attractors from mutual information},
    journal = {Phys. Rev. A},
    year = {1986},
    volume = {33},
    pages = {1134-1140}
}

@article{Kennel1992,
    author = {Matthew B. Kennel and Reggie Brown and Henry D. I, Abarbanel},
    title = {Determining embedding dimension for phase-space reconstruction using a geometrical construction},
    journal = {Phys. Rev. A},
    year = {1992},
    volume = {45},
    pages = {3404-3411}
}

@inproceedings{Takens1981,
    author = {Floris Takens},
    title = {Detecting strange attractors in turbulence},
    booktitle = {Dynamical Systems and Turbulence, Lecture Notes in Mathematics},
    year = {1981},
    volume = {898},
    pages = {366-381}
}

@article{Kligfield2007,
author = {Paul Kligfield  and Leonard S. Gettes  and James J. Bailey  and Rory Childers  and Barbara J. Deal  and E. William Hancock  and Gerard van Herpen  and Jan A. Kors  and Peter Macfarlane  and David M. Mirvis  and Olle Pahlm  and Pentti Rautaharju  and Galen S. Wagner },
title = {Recommendations for the Standardization and Interpretation of the Electrocardiogram},
journal = {Circulation},
volume = {115},
pages = {1306-1324},
year = {2007}}

@article{Butterworth1930,
    author = {Stephen Butterworth},
    title = {On the {T}heory of {F}ilter {A}mplifiers},
    journal = {Exp. Wirel. Wirel. Eng.},
    year = {1930},
    pages = {536-541},
    volume = {7}
}

@article{Teplitzky2020,
    author = {Benjamin A. Teplitzky and Micheal McRoberts and Hamid Ghanbari},
    title = {Deep learning for comprehensive {ECG} annotation},
    journal = {Heart Rhythm},
    year = {2020},
    volume = {17},
    pages = {881-888}
}

@article{Lee2024,
    author = {Jaewon Lee and Miyoung Shin},
    title = {Using beat score maps with successive segmentation for {ECG} classification without {R}-peak detection},
    journal = {Biomed. Signal Process. Control},
    year = {2024},
    volume = {91},
    pages = {105982}
}

@article{Shahi2021,
	title        = {Long-{T}ime {P}rediction of {A}rrhythmic {C}ardiac {A}ction {P}otentials {U}sing {R}ecurrent {N}eural {N}etworks and {R}eservoir {C}omputing},
	author       = {Shahrokh Shahi and Christopher D. Marcotte and Conner J. Herndon and Flavio H. Fenton and Yohannes Shiferaw and  Elizabeth M. Cherry},
	year         = 2021,
	journal      = {Front. Physiol.},
	volume       = 12,
	pages        = 734178
}

@article{Brucke2024,
	title        = {Benchmarking reservoir computing for residential energy demand forecasting},
	author       = {Karoline Brucke and Simon Schmitz and Daniel Köglmayr and Sebastian Baur and Christoph Räth and and Esmail Ansari and Peter Klement},
	year         = {2024},
	journal      = {Energy Build.},
	volume       = {314},
	pages        = {114236}
}

@article{Herteux2024,
	title        = {Forecasting trends in food security with real time data},
	author       = {Joschka Herteux and Christoph Räth and Giulia Martini and Amine Baha and Kyriacos Koupparis and Ilaria Lauzana and Duccio Piovani},
	year         = {2024},
	journal      = {Commun. Earth Environ.},
	volume       = {5},
	pages        = {611}
}

@article{Pathak2017,
    title = {Using machine learning to replicate chaotic attractors and calculate {L}yapunov exponents from data},
    journal = {Chaos},
    volume = {27},
    pages = {121102},
    year = {2017},
    author = {Jaideep Pathak and Zhixin Lu and Brian R. Hunt and Michelle Girvan and Edward Ott}}

@article{Lu2018,
    title = {Attractor reconstruction by machine learning},
    journal = {Chaos},
    volume = {28},
    pages = {061104},
    year = {2018},
    author = {Zhixin Lu and Brian R. Hunt and Edward Ott}}

@article{Watts1998,
    author = {Duncan J. Watts and Steven H. Strogatz},
    title = {Collective dynamics of `small-world' networks},
    journal = {Nature},
    volume = {393},
    year = {1998}
}

@article{Erdos1959,
    author = {Paul Erdős and Alfréd Rényi},
    title = {On random graphs {I}.},
    journal = {Publ. Math. Debr.},
    year = {1959},
    volume = {6},
    pages = {290-297}
}

@article{Herteux2020,
    author = {Joschka Herteux and Christoph Räth},
    title = {Breaking symmetries of the reservoir equations in echo state networks},
    journal = {Chaos},
    year = {2020},
    volume = {30},
    pages = {123142}
}

@article{Bishop1995,
  author  = {Chris M. Bishop},
  title   = {Training with {N}oise is {E}quivalent to {T}ikhonov {R}egularization},
  journal = {Neural Comput.},
  volume  = {7},
  pages   = {108-116},
  year    = {1995},
}

@article{Wikner2024,
    author = {Alexander Wikner and Joseph Harvey and Michelle Girvan and Brian R. Hunt and Andrew Pomerance and Thomas Antonsen and Edward Ott},
    title = {Stabilizing machine learning prediction of dynamics: {N}ovel noise-inspired regularization tested with reservoir computing},
    journal = {Neural Netw.},
    volume = {170},
    year = {2024},
    pages = {94-110}
}

@inproceedings{Lim2021,
 author = {Lim, Soon Hoe and Erichson, N. Benjamin and Hodgkinson, Liam and Mahoney, Michael},
 booktitle = {Advances in Neural Information Processing Systems},
 pages = {5124-5137},
 title = {{N}oisy {R}ecurrent {N}eural {N}etworks},
 volume = {34},
 year = {2021}
}

@article{Smirnov1948,
    author = {Nikolai Smirnov},
    title = {Table for {E}stimating the {G}oodness of {F}it of {E}mpirical {D}istributions},
    journal = {Ann. Math. Statist.},
    year = {1948},
    volume = {19},
    pages = {279-281}
}

@article{Kolmogorov1933,
    author = {Andrey Kolmogorov},
    title = {Sulla determinazione empirica di una legge di distribuzione},
    journal = {G. Ist. Ital. Attuari.},
    year = {1933},
    volume = {4},
    pages = {83-91}
}

@article{Massey1951,
    author = {Frank J. {Massey, Jr.}},
    title = {The {K}olmogorov--{S}mirnov {T}est for {G}oodness of {F}it},
    journal = {JASA},
    year = {1951},
    volume = {46},
    number = {153},
    pages = {68-78}
}

@article{Marsaglia2003,
    author = {George Marsaglia and Wai Wan Tsang and Jingbo Wang},
    title = {{E}valuating {K}olmogorov's {D}istribution},
    journal = {J. Stat. Softw.},
    year = {2003},
    volume = {8},
    issue = {18},
    pages = {1-4}
}

@article{Vrbik2018,
    author = {Jan Vrbik},
    title = {Small-{S}ample {C}orrections to {K}olmogorov--{S}mirnov {T}est {S}tatistic},
    journal = {PJTAS},
    year = {2018},
    volume = {15},
    pages = {15-23}
}

@article{Dormand1980,
    author = {John R. Dormand and Pete J. Prince},
    title = {A family of embedded {R}unge--{K}utta formulae},
    journal = {J. Comput. Appl. Math.},
    year = {1980},
    pages = {19-26},
    volume = {6},
    number = {1}
}

@article{Eckmann1985,
    author = {J.-P. Eckmann and D. Ruelle},
    title = {Ergodic theory of chaos and strange attractors},
    journal = {Rev. Mod. Phys.},
    year = {1985},
    pages = {617-656},
    volume = {57}
}

@article{Grassberger1983,
    author = {Peter Grassberger and Itamar Procaccia},
    title = {Measuring the strangeness of strange attractors},
    journal = {Physica D},
    year = {1983},
    pages = {189-208},
    volume = {9}
}

@article{Gauthier2021,
    author = {Daniel J. Gauthier and Erik Bollt and Aaron Griffith and Wendson A. S. Barbosa},
    title = {Next generation reservoir computing},
    journal = {Nat. Commun.},
    year = {2021},
    volume = {12},
    pages = {5564}
}

@article{Ma2023,
    author = {Haochun Ma and Davide Prosperino and Christoph Räth},
    title = {A novel approach to minimal reservoir computing},
    journal = {Sci. Rep.},
    year = {2023},
    volume = {13},
    pages = {12970}
}

@article{Rosenstein1993,
    author = {Michael T. Rosenstein and James J. Collins and Carlo J. {De Luca}},
    title = {A practical method for calculating largest Lyapunov exponents from small data sets},
    journal = {Physica D},
    year = {1993},
    volume = {65},
    pages = {117-134}
}

@inbook{Lukosevicius2012,
    author={Mantas Lukoševičius},
    title={A {P}ractical {G}uide to {A}pplying {E}cho {S}tate {N}etworks},
    bookTitle={{N}eural {N}etworks: {T}ricks of the {T}rade"},
    year={2012},
    publisher={Springer Berlin Heidelberg},
    address={Berlin, Heidelberg},
    pages={659-686}
}

@article{Racca2021,
    author = {Alverto Racca and Luca Magri},
    title = {Robust {O}ptimization and {V}alidation of {E}cho {S}tate {N}etworks for learning chaotic dynamics},
    journal = {Neural Netw.},
    year = {2021},
    volume = {142},
    pages = {252-268}
}

@inproceedings{Kim2022,
    author = {Siwon Kim and Kukjin Choi and Hyun-Soo Choi and Byunghan Lee and Sungroh Yoon},
    title = {Towards a {R}igorous {E}valuation of {T}ime-{S}eries {A}nomaly {D}etection},
    booktitle = {Proceedings of the Thirty-Sixth AAAI Conference on Artificial Intelligence (AAAI-22)},
    year = {2022},
    pages = {7149-7201}
}

@article{Haluszczynski2019,
    author = {Alexander Haluszczynski and Christoph Räth},
    title = {Good and bad predictions: {A}ssessing and improving the replication of chaotic attractors by means of reservoir computing },
    journal = {Chaos},
    year = {2019},
    volume = {29},
    pages = {103143}
}

@article{Kato2024,
    author = {Junya Kato and Gouhei Tanaka and Ryosho Nakane and Akira Hirose},
    title = {Reconstructive reservoir computing for anomaly detection in time-series signals},
    journal = {Nonlinear Theory Appl. IEICE},
    volume = {15},
    pages = {183-204},
    year = {2024}
}

@article{Wang2022,
    author = {Mingzhao Wang and Zuntao Fu},
    title = {A new method of nonlinear causality detection: {R}eservoir computing {G}ranger causality},
    journal = {Chaos Solitons Fractals},
    year = {2022},
    volume = {154},
    pages = {111675}
}

@article{Zhao2024,
    author = {Jintong Zhao and Zhongxue Gan and Ruixi Huang and Chun Guan and Jifan Shi and Siyang Leng},
    title = {Detecting dynamical causality via intervened reservoir computing},
    journal = {Commun. Phys.},
    year = {2024},
    volume = {7},
    pages = {232}
}

@article{Tamura2026,
    author = {Hiroto Tamura and Kantaro Fujiwara and Kazuyuki Aihara and Gouhei Tanaka},
    title = {Distributional reservoir state analysis for real-time anomaly detection in multivariate time series data},
    journal = {npj Artif. Intell.},
    year = {2026},
    volume = {2},
    pages = {41}
}

@article{Banerjee2019,
    author = {Amitava Banerjee and Jaideep Pathak and Rajarshi Roy and Juan G. Restrepo and Edward Ott},
    title = {Using machine learning to assess short term causal dependence and infer network links},
    journal = {Chaos},
    year = {2019},
    volume = {29},
    pages = {121104}
}

@article{Cao2025,
    title = {Reservoir cross mapping as a nonlinear framework for detecting dynamical causality},
    journal = {Cell Rep. Phys. Sci.},
    volume = {6},
    pages = {102683},
    year = {2025},
    author = {Ren Cao and Jintong Zhao and Chun Guan and Huanfei Ma and Jifan Shi and Siyang Leng}
}

\section*{Author contributions statement}
\textbf{D.P.}: Conceptualization, Methodology, Software, Validation, Investigation, Writing - Original Draft, Writing - Review \& Editing,  Visualization.
\textbf{H.M.}: Conceptualization, Methodology, Writing - Review \& Editing.
\textbf{C.R.}: Conceptualization, Methodology, Writing - Review \& Editing, Supervision.

\section*{Additional information}

\subsection*{Competing interests}
The authors declare no competing interests.

\subsection*{Code availability}
The custom code used to generate and analyze the results in this study is publicly available at \url{https://gitlab.com/dprosperino/unsupervised-anomaly-detection-with-reservoir-computers}.

\subsection*{Data availability}
The synthetic data analyzed in this study were generated numerically.
The model equations, integration scheme, and parameter values needed to reproduce them in full are given in the Methods section.
The specific trajectories used in our experiments are additionally available from the corresponding author on reasonable request.
The clinical electrocardiogram data are publicly available:
We used record 418 of the MIT-BIH Malignant Ventricular Ectopy Database\cite{Greenwald1986}, hosted on PhysioNet\cite{Goldberger2000} and accessible at \url{https://physionet.org/content/vfdb/}.

\end{document}